%% file: hulu_tkde.tex
\documentclass[10pt,journal,compsoc]{IEEEtran}
\ifCLASSOPTIONcompsoc
  \usepackage[nocompress]{cite}
\else
  \usepackage{cite}
\fi
\usepackage{booktabs}
\usepackage{epsfig}
\usepackage{graphicx}
\usepackage{amsmath}
\usepackage{amssymb}

\usepackage{url}
\usepackage{tabularx}
\usepackage{multirow}
\usepackage{subfigure}
\usepackage{amsmath,dsfont}
\usepackage{array}
\usepackage{enumitem}

\usepackage{pifont}

\usepackage{verbatim}

\usepackage{colortbl}
\definecolor{mygray}{gray}{.75}

\usepackage{xcolor}

\usepackage[normalem]{ulem}

\newcommand{\cmark}{\ding{51}}%
\newcommand{\xmark}{\ding{55}}%

\newcommand{\argmin}{\operatornamewithlimits{argmin}}

\newcommand{\eg}{\emph{e.g.,}~}
\newcommand{\etal}{\emph{et al.}~}
\newcommand{\ie}{\emph{i.e.,}~}

\ifCLASSINFOpdf
\else
\fi
\begin{document}
%
\title{Feature Re-Learning with Data Augmentation for Video Relevance Prediction}
%
%
%
%

\author{Jianfeng~Dong*,
        Xun~Wang*,~\IEEEmembership{Member,~IEEE,}
        Leimin~Zhang,
        Chaoxi~Xu,
        Gang~Yang,~
        Xirong~Li
\IEEEcompsocitemizethanks{\IEEEcompsocthanksitem J. Dong, X. Wang and L. Zhang are with the College of Computer and Information Engineering, Zhejiang Gongshang University, Hangzhou 310035, China. * indicates the co-first author. \protect\\
E-mail: dongjf24@gmail.com
\IEEEcompsocthanksitem C. Xu, G. Yang and X. Li are with the Key Lab of Data Engineering and Knowledge Engineering, Renmin University of China, and the AI \& Media Computing Lab, School of Information, Renmin University of China, Beijing 100872, China.\protect\\
E-mail: xirong@ruc.edu.cn
}
\thanks{Manuscript received 31 May. 2019; revised 13 Aug. 2019 and 15 Sep. 2019; accepted 30 Sep. 2019 (Corresponding author: Xirong Li)}}
%
%

\markboth{IEEE Transactions on Knowledge and Data Engineering,~Vol.~x, No.~x, Oct.~2019}%
{Dong \MakeLowercase{\textit{et al.}}: Feature Re-Learning with Data Augmentation for Video Relevance Prediction}
%



\IEEEtitleabstractindextext{%
\begin{abstract}
Predicting the relevance between two given videos with respect to their visual content is a key component for content-based video recommendation and retrieval. Thanks to the increasing availability of pre-trained image and video convolutional neural network models, deep visual features are widely used for video content representation. However, as how two videos are relevant is task-dependent, such off-the-shelf features are not always optimal for all tasks. Moreover, due to varied concerns including copyright, privacy and security, one might have access to only pre-computed video features rather than original videos. We propose in this paper feature re-learning for improving video relevance prediction, with no need of revisiting the original video content. In particular, re-learning is realized by projecting a given deep feature into a new space by an affine transformation. We optimize the re-learning process by a novel negative-enhanced triplet ranking loss. In order to generate more training data, we propose a new data augmentation strategy which works directly on frame-level and video-level features. Extensive experiments in the context of the Hulu Content-based Video Relevance Prediction Challenge 2018 justify the effectiveness of the proposed method and its state-of-the-art performance for content-based video relevance prediction.
\end{abstract}
}

\maketitle

\IEEEdisplaynontitleabstractindextext

%
\IEEEpeerreviewmaketitle

\IEEEraisesectionheading{\section{Introduction}\label{sec:introduction}}

%
%
%
%

\input{intro}

\section{Related Work} \label{sec:rel-work}

\input{rel-work}

\section{Proposed Video Relevance Prediction} \label{sec:solution}

\input{solution}

\section{Evaluation} \label{sec:eval}

\input{eval}

\section{Conclusions} \label{sec:conc}

To predict task-specific video relevance, this paper proposes a ranking-oriented feature re-learning model with feature-level data augmentation. The model is trained with a novel negative-enhanced triplet ranking loss (NETRL) on the ground-truth data with respect to a given task.
Extensive experiments on real-world datasets provided by the HULU Content-based Relevance Video Relevance Prediction Challenge support the following conclusions.
Compared with the commonly used triplet ranking loss, NETRL not only improves the performance but only shows faster convergence.
If the relationship of a candidate video with respect to another candidate video is available, such information can be exploited to improve video relevance prediction. 
The proposed multi-level data augmentation strikes a good balance between the model's effectiveness and its robustness with respect to Gaussian noises. While the evaluation is conducted in the context of the HULU challenge, we believe the proposed method also has a potential for other tasks that require content-based video relevance prediction.


%



\ifCLASSOPTIONcompsoc
  \section*{Acknowledgments}
\else
  \section*{Acknowledgment}
\fi

This work was supported by the Zhejiang Provincial Natural Science Foundation (No. LQ19F020002), the National Natural Science Foundation of China (No. U1609215, No. 61672523, No. 61902347), and the Fundamental Research Funds for the Central Universities and the Research Funds of Renmin University of China (No. 18XNLG19).
The authors are grateful to the Hulu Content-based Video Relevance Prediction Challenge organizers (Dr. Xiaohui Xie and Dr. Peng Wang) for evaluating their results on the test of TV-shows and Movies. The authors also thank the anonymous reviewers for their insightful comments.

\ifCLASSOPTIONcaptionsoff
  \newpage
\fi



%



\bibliographystyle{IEEEtran}
\bibliography{hulu}

%

\begin{IEEEbiography}[{\includegraphics[width=1in,height=1.25in,clip,keepaspectratio]{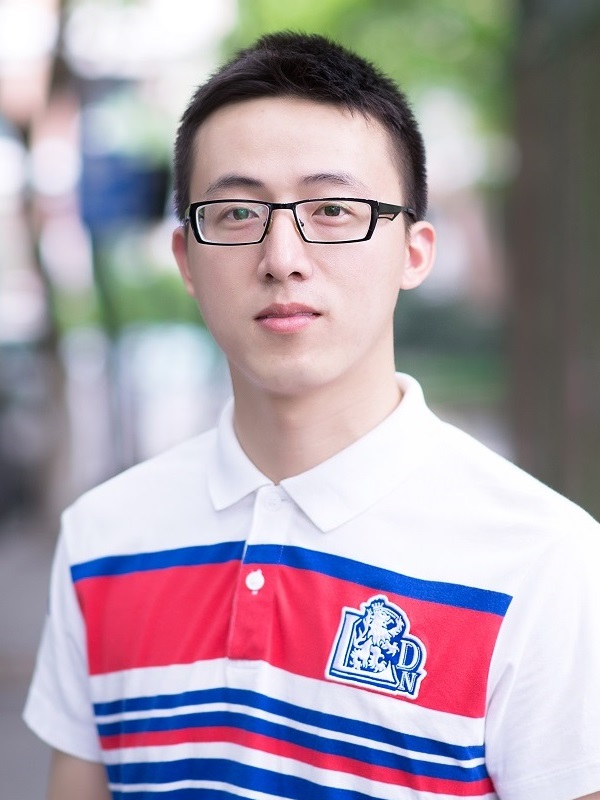}}]
{Jianfeng Dong} received the B.E. degree in software engineering from Zhejiang University of Technology in 2009, and the Ph.D. degree in computer science from Zhejiang University in 2018, all in Hangzhou, China. He is currently an Assistant Professor at the College of Computer and Information Engineering, Zhejiang Gongshang University, Hangzhou, China. 

His research interests include multimedia understanding, retrieval and recommendation. He was awarded the ACM Multimedia Grand Challenge Award in 2016.
\end{IEEEbiography}

\begin{IEEEbiography}[{\includegraphics[width=1in,height=1.25in,clip,keepaspectratio]{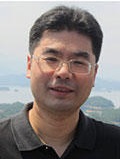}}]
{Xun Wang} is currently a professor at the School of Computer Science and Information Engineering, Zhejiang Gongshang University, China. He
received his BSc in mechanics, Ph.D. degrees in computer science, all from Zhejiang University, Hangzhou, China, in 1990 and 2006, respectively. His current research interests include mobile graphics computing, image/video processing, pattern recognition, intelligent information processing and visualization. He is a member of the IEEE and ACM, and a senior member of CCF.
\end{IEEEbiography}

\begin{IEEEbiography}[{\includegraphics[width=1in,height=1.25in,clip,keepaspectratio]{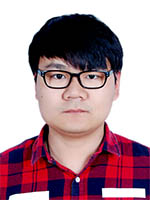}}]
{Leimin Zhang} received his B.S. degree in Computer Science and Technology from Zhejiang Sci-Tech University in 2017. He is currently a graduate student at the College of Computer and Information Engineering, Zhejiang Gongshang University, pursuing his master degree on multimedia recommendation.
\end{IEEEbiography}

\begin{IEEEbiography}[{\includegraphics[width=1in,height=1.25in,clip,keepaspectratio]{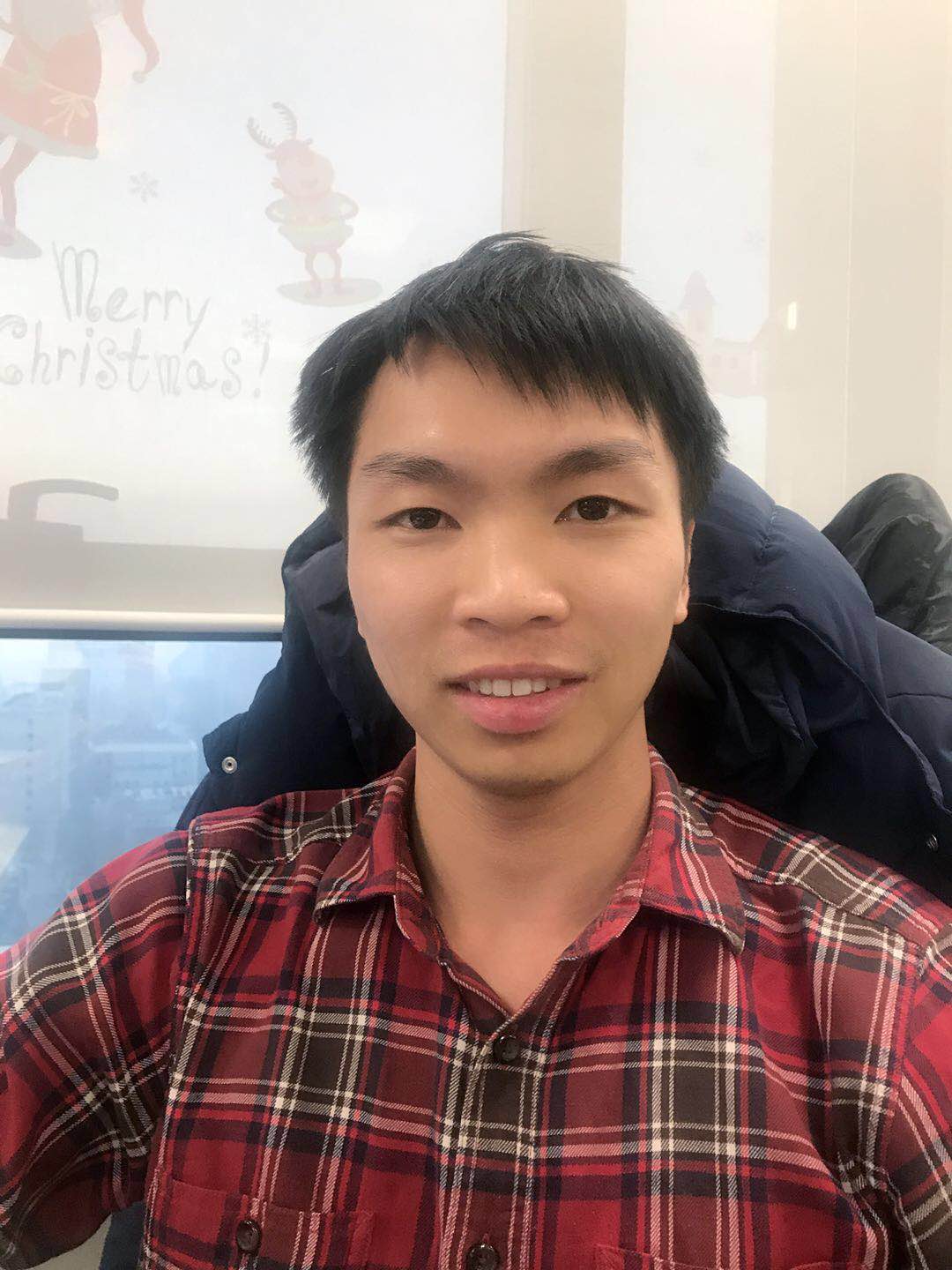}}]
{Chaoxi Xu} received his B.S. degree in Computer Science from Renmin University of China, Beijing, China in 2017. He is currently a graduate student at School of Information, Renmin University of China, pursuing his master degree on cross-lingual multimedia retrieval.
\end{IEEEbiography}

\vspace{-100 mm}

\begin{IEEEbiography}[{\includegraphics[width=1in,height=1.25in,clip,keepaspectratio]{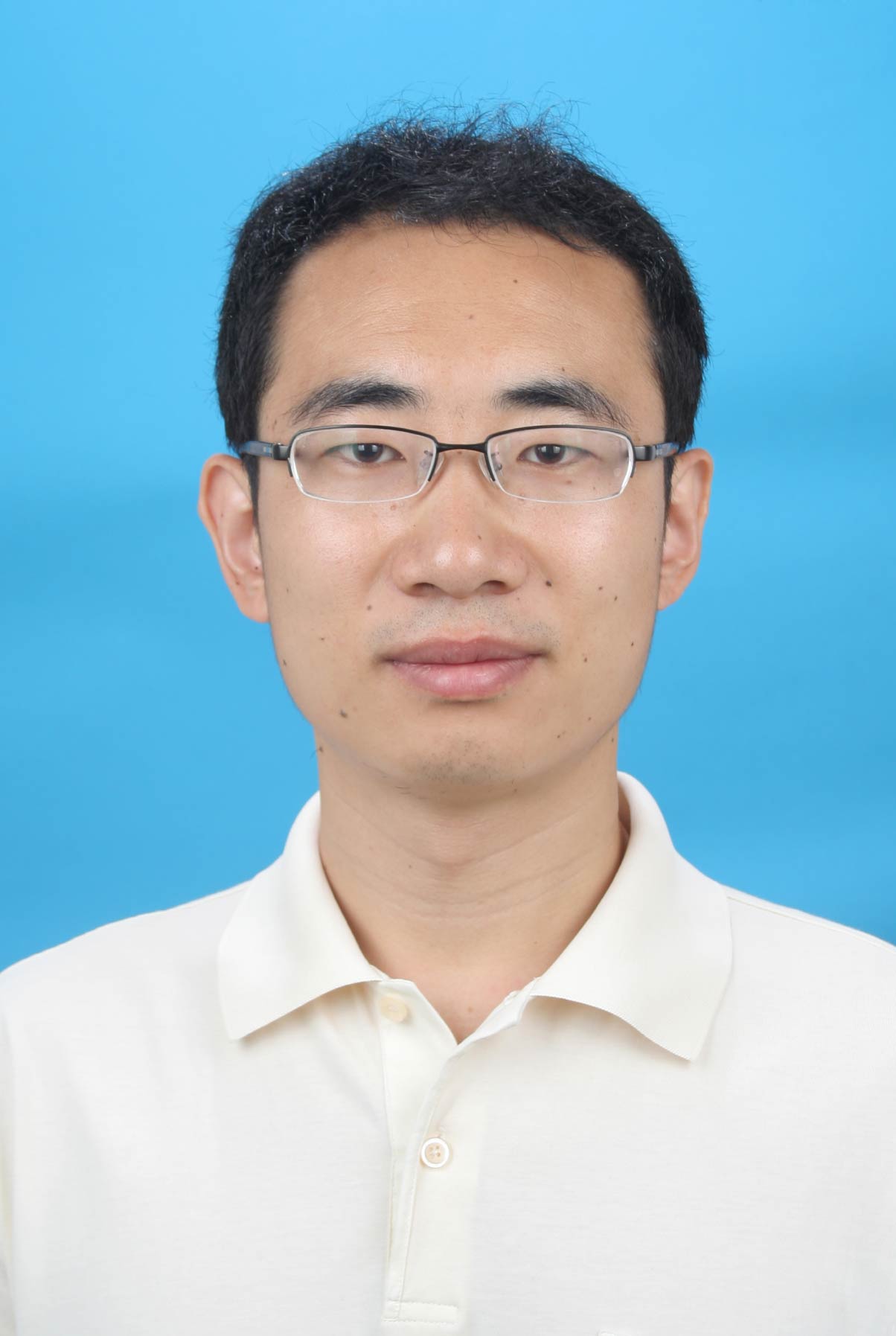}}]
{Gang Yang} received his Ph.D. degree in Innovative Life Science from University of Toyama, Toyama, Japan in 2009. He is currently an Associate Professor at School of Information, Renmin University of China, Beijing, China. His current research interests include computational intelligence, multimedia computing and machine learning.
\end{IEEEbiography}

\vspace{-100 mm}

\begin{IEEEbiography}[{\includegraphics[width=1in,height=1.25in,clip,keepaspectratio]{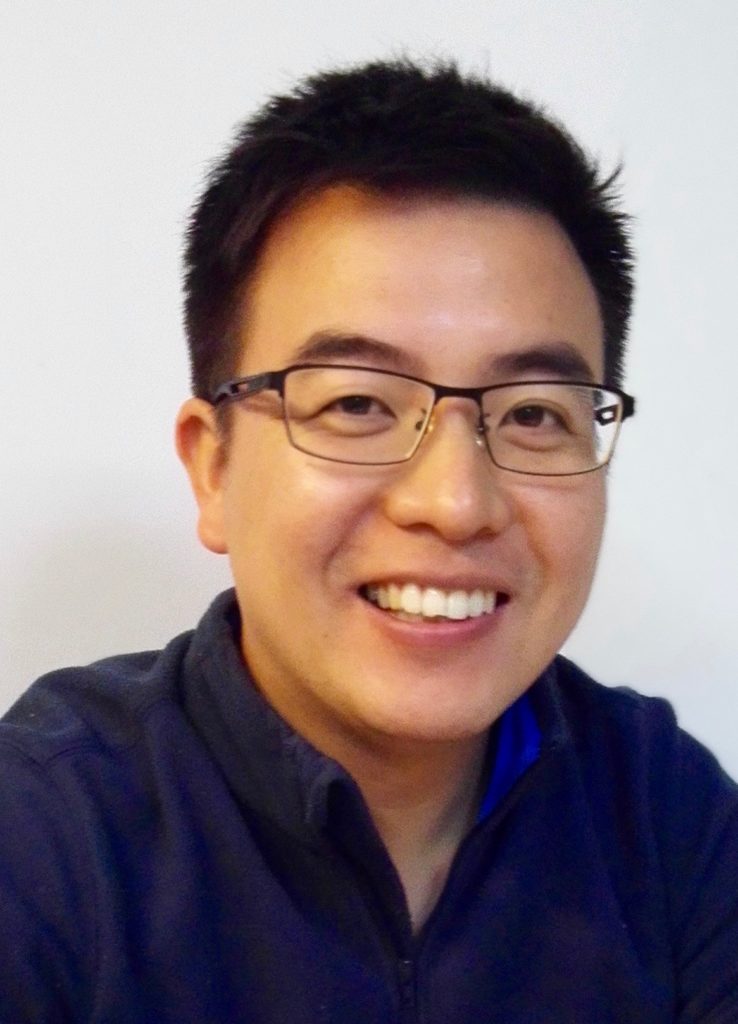}}]
{Xirong Li} received the B.S. and M.E. degrees from Tsinghua University, Beijing, China, in 2005 and 2007, respectively, and the Ph.D. degree from the University of Amsterdam, Amsterdam, The Netherlands, in 2012, all in computer science. He is currently an Associate Professor with the Key Lab of Data Engineering and Knowledge Engineering, Renmin University of China, Beijing, China. His research includes image and video retrieval.

Dr. Li was recipient of the ACMMM 2016 Grand Challenge Award, the ACM SIGMM Best Ph.D. Thesis Award 2013, the IEEE TRANSACTIONS ON MULTIMEDIA Prize Paper Award 2012, and the Best Paper Award of ACM CIVR 2010. He was area chair of ACMMM 2019 / 2018 and ICPR 2016.

\end{IEEEbiography}








\end{document}

%% file: intro.tex
\IEEEPARstart{P}{redicting} the relevance between two given videos is essential for a number of video-related tasks including video recommendation \cite{hulu-baseline,he2018nais}, video annotation \cite{tip16-graph-learning,wang2012event}, category video retrieval \cite{liu2017deep}, near-duplicate video retrieval \cite{liu2013near}, video copy detection \cite{liu2012segmentation} and so on. 
In the context of video recommendation, a recommendation system aims to suggest videos which may be of interest to a specific user. To that end, the video relevance shall reflect the user's (implicit) feedback such as watch, search and browsing history. For category video retrieval, one wants to search for videos that are semantically similar to a given query video. As such, the video relevance should reflect the semantical similarity. As for near-duplicate video retrieval, one would like to retrieve videos showing exactly the same story, but with minor photographic differences and editions with respect to a given query video. For this purpose, the video relevance shall reflect the visual similarity. It is clear that the optimal approach to video relevance prediction is task dependent.

\begin{figure}[tb!]
\centering
 \subfigure [Off-the-shelf feature space] {
\noindent\includegraphics[width=0.6\columnwidth]{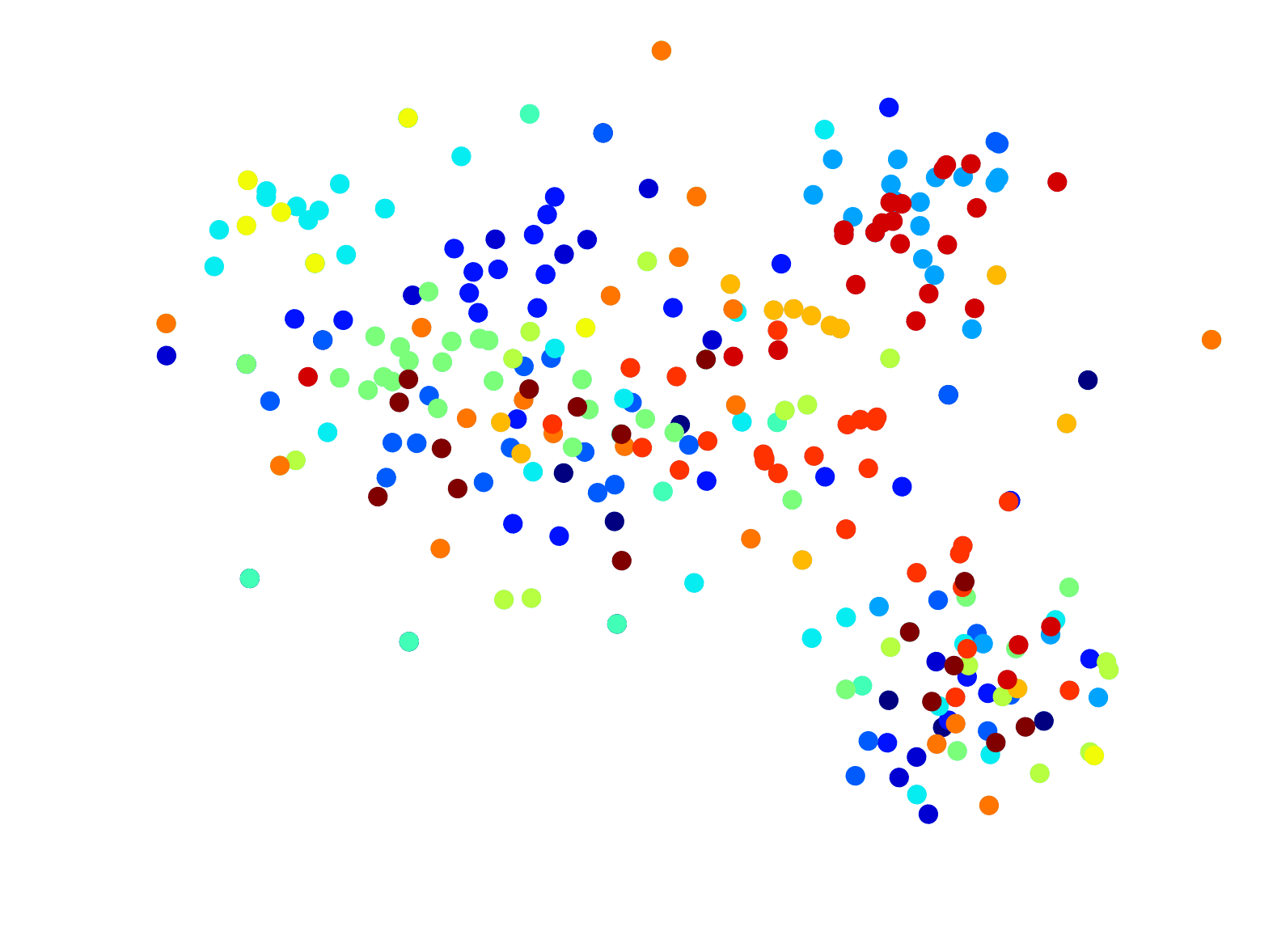}
\label{fig:original_space}}
 \subfigure[Feature space re-learned by this work] {
\noindent\includegraphics[width=0.65\columnwidth]{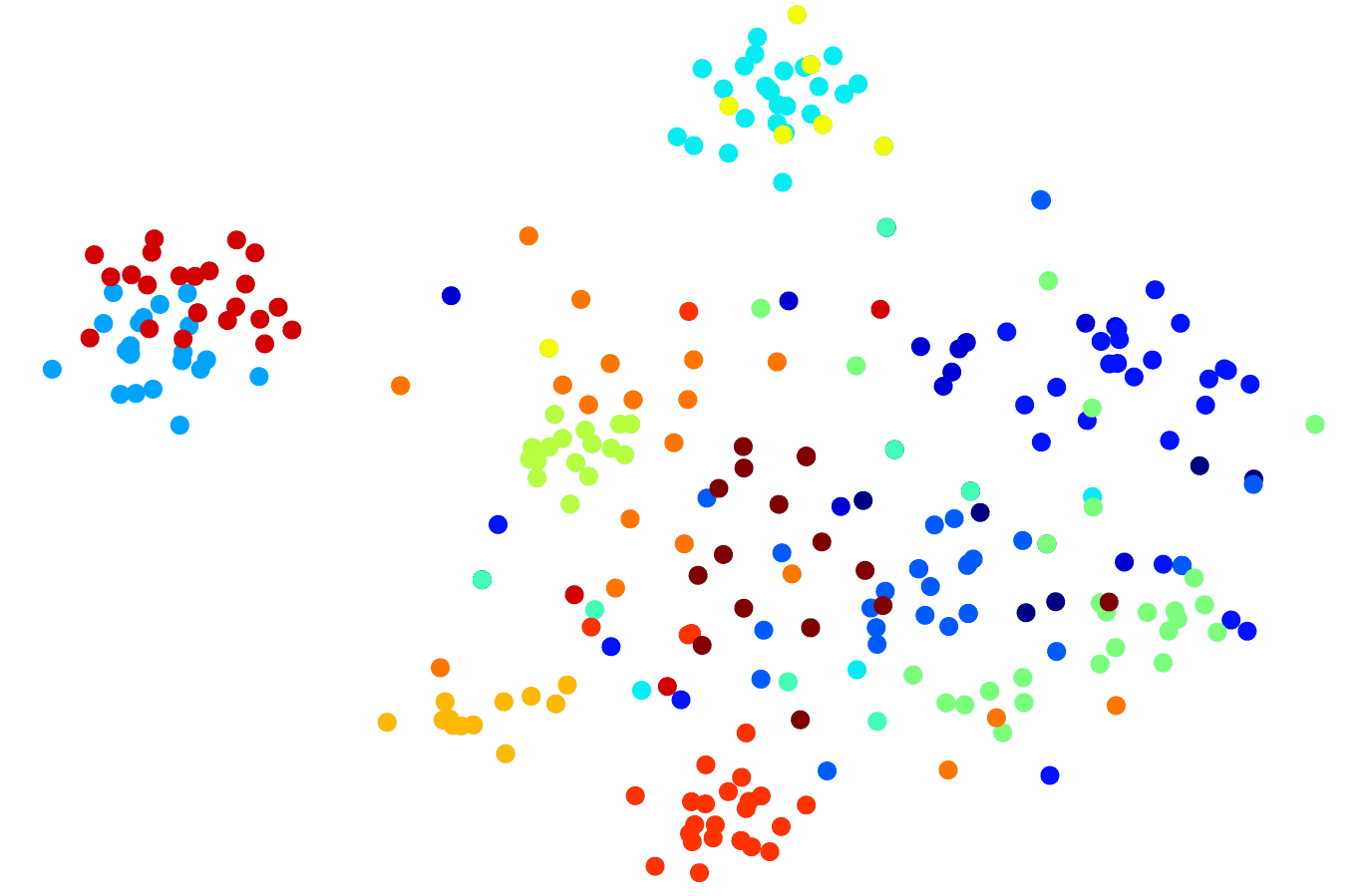}
\label{fig:re_learned_space}}
\caption{Off-the-shelf feature space versus re-learned feature space. In the context of video recommendation, we randomly select 15 query videos and their corresponding relevant videos from the validation set of the TV-shows dataset \cite{hulu-baseline}, and use t-SNE \cite{maaten2008tsne} to visualize their distribution in (a) the off-the-shelf feature space obtained by pre-trained CNN model for semantic classification and (b) the re-learned feature space obtained by our proposed model. 
Dots with the same color indicate videos relevant to a specific query. The plots reveal that relevant videos stay closer in the re-learned feature space than in the off-the-shelf feature space. Off-the-shelf feature: Inception-v3. Best viewed in color. }
 \label{fig:tsne_show}
\end{figure}

In order to estimate video relevance, some works \cite{basu1998recommendation,mei2007videoreach} utilize textual content of videos. For instance, Basu \etal \cite{basu1998recommendation} utilize meta data associated with videos, such as titles, keywords, and director names.
However, meta data is not always available and its quality is not guaranteed, especially for user-generated videos. For example, a title is easily alterable, which may be deliberately written to attract users while irrelevant to the video content itself \cite{du2018personalized}. Hence, video relevance prediction approaches depending on the textual content may lead to poor performance. In contrast to the textual content, a video's visual content is available right after the video is created and more reliable. In this work, we focus on the visual content to predict the relevance between two videos.

Existing works largely use off-the-shelf visual features to predict video relevance \cite{yang2007online,deldjoo2018audio,chivadshetti2015content,kordopatis2017near,wang2012movie2comics}. In an earlier work on video recommendation, Yang \etal \cite{yang2007online} estimate video relevance by the Manhattan distance in terms of color, motion intensity and shot frequency features. A recent work by Deldjoo \etal \cite{deldjoo2018audio} uses a more advanced feature extracted by a pre-trained Convolutional Neural Network (CNN) model, considering its superior performance in multiple vision related tasks \cite{xie2019convolutional,miccai19-cataract,xie2019automated,chen2019structure,wang2017first}.
%
Relevance prediction using off-the-shelf visual features is also common in other video relevance related tasks. For category video retrieval, Chivadshetti \etal \cite{chivadshetti2015content} utilize color, texture and motion features to represent videos, and the relevance between two videos is measured as the similarity of their corresponding features. 
For near-duplicate video retrieval, Kordopatis-Zilos \etal \cite{kordopatis2017near} extract multiple intermediate features from video frames by a pre-trained CNN model. Their video relevance is then computed as the cosine similarity between video-level features obtained by aggregating multiple features with bag-of-visual-words encoding.
In general, the off-the-shelf features are not tailored to the needs of a specific task. Consequently, video relevance prediction using such features tends to be suboptimal.
As Fig. \ref{fig:original_space} shows, in the context of visual content based video recommendation, relevant videos (denoted by the same colored dots) tend to scatter in the off-the-shelf video feature space.

Note that video relevance prediction is orthogonal to traditional video similarity calculation. The latter uses as is a provided feature, let it be low-level motion features or high-level semantic vectors \cite{tmm16-tagbook}, with emphasis on improving efficiency by dimension reduction \cite{huang2007dual} or hashing techniques \cite{shen2018unsupervised,song2013effective}. In principle, such techniques can be leveraged for accelerating video relevance prediction if efficiency is in demand.

We note some initial efforts for  learning a new video feature space and consequently measuring the video relevance in the new space \cite{hulu-baseline,fusedlstm,kordopatis2017near2,lee2017large}. 
The above process transforms a given feature into a new space that has a better discrimination ability for video relevance prediction. We coin this \emph{feature re-learning}. The essential difference between feature re-learning and traditional feature transform is two-fold. First, the prefix ``re-'' emphasizes the given feature is a (deeply) learned representation. By contrast, the given feature in a traditional setting is typically low-level, \eg bag of local descriptors. Improving over an already learned feature is more challenging. Consequently, supervised learning is a must for feature re-learning. By contrast, traditional feature transformation can be unsupervised, \eg Principle Component Analysis or random projection.

For learning based methods, the choice of the loss function is important. The previous works utilize a triplet ranking loss \cite{chechik2010large} which preserves the relative similarity among videos. The loss needs relevant video pairs for training and aims to make the similarity between relevant video pairs larger than that between irrelevant pairs in the learned feature space.  Note that the triplet ranking loss, as focusing on the relative distance, ignores how close (or how far) between the relevant (or irrelevant) video pairs, which affects its effectiveness for training a good model. 
In this work, we propose a novel negative-enhanced triplet ranking loss (NETRL) that effectively considers both relative and absolute similarity among videos.

It is well recognized that the more data a learning based method has access to, the more effective it can be. However, collecting a large amount of relevant video pairs is both time-consuming and expensive. 
Moreover, as more original data is exposed for training, it is more likely to cause privacy and data security issues \cite{larson2017towards}. These issues can be serious, as videos such as TV-shows or movies are typically copyright-protected. 
One way to relieve these issues is data augmentation, which generates more training samples based on existing data, and thus less original data is required. However, common data augmentation strategies such as flipping, rotation, zooming in/out, have to be conducted on the video content. With the requirement of revisiting the original video content, the privacy and the security issues remain. We develop a new data augmentation strategy that works directly with fame-level and video-level features, with no need of re-accessing original videos.

In this paper, we study the video relevance prediction in the context of the Hulu Content-based Video Relevance Prediction Challenge \cite{hulu-baseline}. In this challenge, given a seed video without any meta data, participants are asked to recommend a list of relevant videos with respect to the given seed video from a set of pre-specified candidate videos. The key of the challenge is to predict relevance between a seed video and a candidate video.
Notice that we as participants have no access to original video data. Instead, the organizers provide two visual features, extracted from individual frames and frame sequences by pre-trained Inception-v3 \cite{abu2016youtube} and C3D \cite{tran2015learning} models, respectively. This challenging scenario is suitable for evaluating our proposed method.
Our contributions are as follows:
\begin{itemize}
    \item We propose a ranking-oriented feature re-learning model with a negative-enhanced triplet ranking (NETR) loss function for video feature prediction. Compared with the commonly used triplet ranking loss, the proposed loss not only gives better performance but also shows faster convergence.
	\item We improve feature re-learning by proposing a new data augmentation strategy. The proposed strategy can be flexibly applied to frame-level or video-level features. Without the need of re-visiting any original videos, our strategy is beneficial for the security protection of video data.
    \item Accompanied with the NERT loss and the feature-level data augmentation, the proposed video feature re-learning solution achieves state-of-the-art performance on two real-world datasets provided by HULU.   
\end{itemize}

A preliminary version of this work was published as a technical note at ACMMM 2018 \cite{ourmm18}, which describes our winning entry for the Hulu Content-based Video Relevance Prediction Challenge \cite{hulu-baseline}. 
In this work, we improve over the conference paper in multiple aspects. First, we introduce a new loss function to supervise the feature re-learning process. Second, for the scenario wherein the relevance relationship between candidate videos is (partially) known, we propose a new formula for video relevance computation. Third, we provide a number of detailed evaluations with respect to the choice of the feature projection architecture, the robustness of the proposed method and the efficiency of applying the trained model for video relevance prediction. All this is not present in the conference edition. Lastly, compared to the best run of the conference paper which is based on model ensemble, the new technical improvements allow us to obtain a new state-of-the-art on the test set with a single model ($0.200$ versus $0.178$ on TV-shows and $0.171$ versus $0.151$ on Movies in terms of recall@100). Data and code are available at \url{https://github.com/danieljf24/cbvr}.

%% file: rel-work.tex
\subsection{Video Relevance Learning}

%
To predict the video relevance, a number of works \cite{hulu-baseline,lee2017large,fusedlstm,dong2018video,kordopatis2017near2,lee2018collaborative} are proposed to re-learn a new video feature space, thus measure video relevance in the learned space by standard distance metric, \eg cosine distance \cite{hulu-baseline,lee2017large}, Euclidean distance\cite{dong2018video}.
For instance, Liu \etal \cite{hulu-baseline} first extract visual features of video frames by pre-trained CNN models and then obtain the video-level feature by mean pooling. A fully connected layer is further employed to map videos into a new feature space and a triplet ranking loss which preserves the relative similarity of videos is used for model training. 
Lee \etal \cite{lee2017large} also utilize fully connected layers and triplet ranking loss for feature re-learning, but both visual and audio features are employed. 
Similar in spirit with \cite{hulu-baseline}, the work by Bhalgat \etal \cite{fusedlstm} utilizes Long Short-Term Memory to exploit the temporal information of videos instead of mean pooling for visual feature aggregation.
For the task of near-duplicate video retrieval, Kordopatis \etal \cite{kordopatis2017near2} employ a 3-layer multilayer perceptron to map videos into a new feature space, and use the same triplet ranking loss with \cite{hulu-baseline} for model training.
For category video retrieval, Dong \etal \cite{dong2018video} fine-tune a CNN model \cite{krizhevsky2012imagenet} to learn a video feature space. Besides the triplet ranking loss, they additionally integrate a classification loss to preserve the semantic similarity of videos in the new feature space. But it needs additional classification ground-truth data for training.
Instead of learning a new feature space for video relevance prediction, Chen \etal \cite{chen2018content} devise a CNN based classification model to determine whether an input video pair is related and the predicted probability is deemed as the relevance score of the video pair.
However, at run time the model by \cite{chen2018content} requires a video to be paired with another video as the network input. By contrast, the models of feature re-learning represent videos in the new space independently, meaning the representation can be precomputed. 
It is an advantageous property for large-scale retrieval applications \cite{xie2018double}.

Our work also aims to learn a new feature space that better reflects task-specific video relevance. For feature re-learning, we propose a novel negative-enhanced triplet ranking loss function which considers both relative and absolute similarity among videos. Moreover, different with \cite{dong2018video} using additionally classification loss, our proposed loss with no need for additional classification ground-truth data.

\subsection{Triplet ranking Loss}
Our proposed loss is rooted from triplet ranking loss, which has been  widely used in many ranking-oriented tasks \cite{frome2013devise,karpathy2015deep,tmm207-dong,wang2017adversarial}. The standard triplet ranking loss (TRL), introduced by Chechik \etal \cite{chechik2010large}, considers a relative distance constraint to encourage the distance of a negative pair larger than the distance of a positive pair with a given margin.  Faghri \etal \cite{faghri2017vse} improve TRL by hard negative mining, where a positive instance is paired with the negative instance most similar to the positive instead of a negative instance sampled at random. 
The contrastive loss by Hadsell \etal \cite{hadsell2006dimensionality} considers the absolute distance, minimizing the distance of the positive pairs, while at the same time enforcing the distance of the negative pairs to be larger than a given margin. 
Although these losses have demonstrated promising performance in the context of image-to-image retrieval \cite{chechik2010large}, cross-model retrieval\cite{faghri2017vse} and handwritten digit recognition \cite{hadsell2006dimensionality}, their effectiveness for video relevance prediction has not been justified. This work resolves this uncertainty. Moreover, we introduce a new loss that effectively exploits both the relative and absolute distance constraint.

\subsection{Data Augmentation}

Data augmentation is a widely used technique in data-driven methods, which aims to create novel samples thus increase the amount of training data.
Existing data augmentation methods are instance-level augmentation, which operate on original instances, such as images and videos, for augmentation.
The traditional practice \cite{bloice2017augmentor,krizhevsky2012imagenet} for augmenting images is to perform affine transformation over the original image, such as cropping, flipping, rotating and zooming in/out the image. In addition to the affine transformation, perspective transformation and color transformation are also applied for image augmentation. For example, given an image, we can change its brightness, contrast or saturation to generate more diverse training images.
Instead of adjusting the instance for augmentation, recent works \cite{perez2017effectiveness,zheng2017unlabeled,zhu2018emotion} propose to generate novel instances for data augmentation. Most of them depend on a generative model, called Generative Adversarial Networks \cite{goodfellow2014generative}, which has a potential of generating novel instances according to the original instance distribution, such as images. For instance, Zheng \etal \cite{zheng2017unlabeled} adopt a generative adversarial network variant \cite{radford2015unsupervised} to additionally generate novel pedestrian images for training. 
For videos, the above augmentation strategies are also applicable, which can be employed on video frames.
In order to augment video data for video classification, Karpathy \etal \cite{karpathy2014large} crop and flip all video frames.
Similarly, Bojarski \etal \cite{bojarski2016end} artificially shift and rotation video frames for the task of self-driving cars.

Different from the above augmentation methods, our proposed augmentation method works for video features instead of original videos. As our method requires no access to original videos, it naturally ensures the security and privacy issues of video data. Moreover, our method is applicable for both frame-level and video-level features. 

Our frame-level data augmentation resembles to some extent segment-based feature extraction commonly used in video data processing. There, a video is first split into several consecutive segments, with segment-level features obtained by averaging features of the frames in each segment. By contrast, segments obtained by our method are not consecutive. Each segment consists of uniformly down-sampled frames, thus better representing the entire video.

Our video-level data augmentation appears to be similar to input noise-injection used in conditional generative models (CGMs) such as GAN \cite{goodfellow2014generative} and CVAE \cite{sohn2015learning}. The major difference is that noise used in CGMs are fed into a generative network to generate diverse samples, while in our work, noises are introduced to improve a discriminative network. Due to this difference, CGMs typically concatenate noise with the input, see \cite{sohn2015learning}. By contrast, our method directly adds noise to the input.

%% file: solution.tex
\begin{figure*}[tb!]
\centering
\includegraphics[width=2\columnwidth]{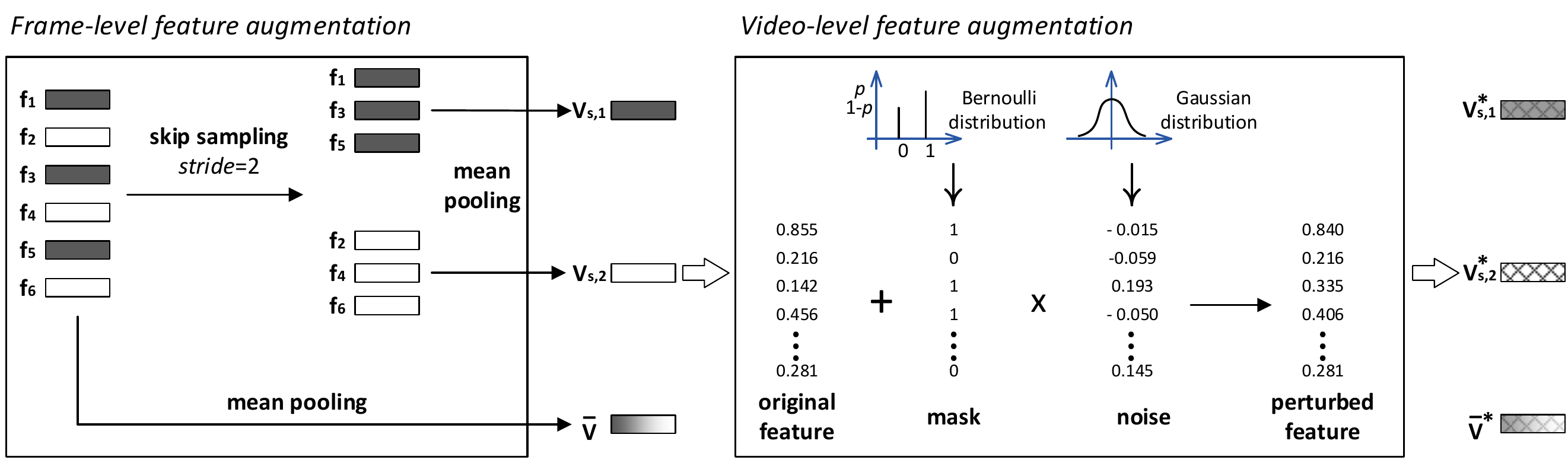}
\caption{Multi-level augmentation strategy for features. It consists of two steps: frame-level feature augmentation and video-level feature augmentation. Given a video with extracted frame-level features, it yields $s+1$ new training instances by multi-level augmentation strategy with skip sampling of a stride $s=2$ for the subsequent supervised learning.}
\label{fig:multi_level_da}
\end{figure*}

We consider \emph{content-based} video relevance prediction. 
Given a video, we use $v \in \mathbb{R}^d$ to indicate a $d$-dimensional deep feature vector that describes its visual content. 
Given two videos $v$ and $v'$, their content-based relevance is typically computed in terms of the cosine similarity between the corresponding features, \ie
\begin{equation} \label{eq:cs-raw}
cs(v,v') = \frac{v \cdot v'}{||v||\cdot ||v'||}.
\end{equation}
As mentioned in Section \ref{sec:introduction}, an off-the-shelf video feature, as extracted by a pre-trained CNN model, does not necessarily lend itself to a specific task. Therefore, we propose to learn a new video feature space, expressed as $\phi(v) \in \mathbb{R}^p$, such that $cs(\phi(v), \phi(v'))$ better reflects task-specific video relevance. In particular, the task-specific property is ensured by learning from the ground-truth data with respect to a given task. 
The ground-truth data is a collection of relevant video pairs, denoted as $\mathcal{D} = \{(v, v^+)\}$.

In what follows, we first introduce ranking-oriented feature re-learning method which maps an off-the-shelf video feature into a new feature space, followed by our proposed multi-level augmentation strategy which considers both the frame-level and video-level features. 
Finally, we present two strategies to predict video relevance in the re-learned video feature space. 
For the ease of reference, main notations defined in this work are listed in Table \ref{tab:notation}.

\subsection{Ranking-oriented Feature Re-learning} \label{ssec:model}

We propose a ranking-oriented feature re-learning method to map videos into a new feature space where relevant videos are near and irrelevant videos are far away. 
Before feeding videos to the feature re-learning model, we choose to first represent each video as a video-level feature vector.
As the number of frame features varies over videos, we employ mean pooling which is simple yet consistently found to be effective in multiple content-based tasks \cite{tmm16-tagbook,mm2016-early,pan2016jointly,hulu-baseline,dong2018predicting}. Note more advanced feature aggregation methods \cite{cvpr2019-dual-dong,arandjelovic2016netvlad} can also be used here.
We then utilize an affine transformation to project an off-the-shelf video feature into a new feature space with the dimensionality of $p$.
More formally, the new feature vector is represented as: 
\begin{equation} \label{eq:affine-transform}
\phi(v) = Wv+b,
\end{equation}
where $W \in \mathbb{R}^{p \times d}$ is trainable affine matrix and $b \in \mathbb{R}^{p}$ indicates a bias term. 
Therefore, given a video pair of ($v$,$v'$), their similarity is estimated as the cosine similarity in terms of their corresponding new video features. 
Note that the affine transformation can be viewed as a one-layer fully connected network (FCN). In principle, a deeper FCN, \eg two-layer FCN or two-layer residual FCN, can also be used. We will investigate which network architecture is most suited for feature re-learning.

\input{table-notations}

In order to train the model, we introduce a negative-enhanced triple ranking loss function. So we firstly describe the triplet ranking loss (TRL), followed by the description of our proposed loss function.

The TRL is widely used and works well in many ranking based tasks \cite{Schroff2015FaceNet,frome2013devise,karpathy2015deep,tmm207-dong}.
As the loss needs triplets for training, we construct a large set of triplets $\mathcal{T} = \{(v, v^+, v^-)\}$ from the relevant video pair set $\mathcal{D}$, where positive $v^+$ and negative $v^-$ indicate videos relevant and irrelevant with respect to video $v$ respectively. The negative video $v^-$ is randomly sampled from training videos. 
Given a triplet of $(v, v^+, v^-)$, the TRL for the given triplet is defined as follows:
\begin{equation}
\begin{array}{r}\label{eq:triplet_loss}
\mathcal{L}^{'}(v, v^+, v^-; W, b) =  \max(0, m_1 - cs_\phi(v,v^+) + cs_\phi(v, v^-)) \\
\end{array}
\end{equation}
where $cs_\phi(v, v')$ denotes the cosine similarity score between $\phi(v)$ and $\phi(v')$, and $m_1$ represents the margin. 
The TRL considers the relative similarity among the triplets, making the similarity between relevant video pairs larger than that between irrelevant video pairs by a constant margin. 
However, we observe that this loss ignores how close (or how far) between the relevant (or irrelevant) video pair in the re-learned feature space, which affects its performance for training a good model for video relevance prediction.

Therefore, we improve the TRL by adding an extra constraint to control the absolute similarity among video pairs.
We add a constraint of negative pairs to the TRL.
The constraint is designed to push negative video pair apart in the re-learned feature space. 
We implement the idea by $max(0, cs_\phi(v, v^-) - m_2)$,  which enforces the similarity of a negative video pair smaller than a given constant $m_2$. By definition, $m_2$ has to be smaller than the maximum of the cosine similarity, which is $1$. 
In the training process, when the similarity of the negative pair in the re-learned feature space is larger than $m_2$, the constraint term will penalize the model to adjust the feature space to make the pair far away. 
By combining TRL and the negative constraint, we develop a new loss termed Negative-Enhanced Triplet Ranking Loss (NETRL), computed as
\begin{equation}
\begin{split}\label{eq:ner_loss}
\mathcal{L}(v, v^+, v^-; W, b) &=  \max(0, m_1 - cs_\phi(v,v^+) + cs_\phi(v, v^-)) \\
&+ \alpha \max(0, cs_\phi(v, v^-) - m_2 ),
\end{split}
\end{equation}
where $\alpha \geqslant 0$ is a trade-off parameter.

One might consider adding a similar constraint on positive pairs, \ie $\max(0, cs_\phi(v,v^+)-m_2)$, as an alternative to the negative constraint. We opt for the latter, because the amount of negative pairs we can learn from is much larger than that of positive pairs. Indeed, the advantage of the negative constraint is confirmed by our experiments.

Finally, we train the feature re-learning model by minimizing the proposed negative-enhanced triplet ranking loss on a triplet set $\mathcal{T} = \{(v, v^+, v^-)\}$, and the overall objective function of the model is as:
\begin{equation}\label{eq:obj}
\argmin_{W, b} \sum_{(v, v^+, v^-)\in \mathcal{T}}  \mathcal{L}(v, v^+, v^-; W, b).
\end{equation}

\subsection{Multi-level Feature Augmentation}  \label{ssec:da}

Data augmentation is one of the effective ways to improve the performance of learning based models, especially when the training data are inadequate.
As mentioned in Section \ref{sec:introduction}, the Hulu organizers do not provide original videos while provide pre-computed video features.
The unavailability of video data means traditional instance-level data augmentation strategies such as flipping, rotating, zooming in/out, are inapplicable. 
Therefore, we introduce a multi-level augmentation strategy that works for video features, with no need for original videos.
As the proposed augmentation strategy performs on video features instead of original videos, which additionally benefits the security protection of video data.
The multi-level augmentation strategy has two steps: frame-level feature augmentation and video-level feature augmentation. Figure \ref{fig:multi_level_da} demonstrates the overview of multi-level augmentation strategy.

\textbf{Frame-level feature augmentation.}
Inspired by the fact that humans could grasp the video topic after watching only several sampled video frames in order, we first augment data by skip sampling.
Given a video of $n$ frames, let $f_i$ be the feature vector of the $i$-th frame. We perform skip sampling with a stride of $s$ over the frame sequence. 
In this way, $s$ new sequences of frame-level features are generated. Accordingly, mean pooling is employed to obtain $s$ new features at the video level, that is
\begin{equation}
\begin{array}{l}
v_{s,1} = \mbox{mean-pooling} \{f_{1}, f_{1+s}, f_{1+2s}, ...\}, \\
v_{s,2} = \mbox{mean-pooling} \{f_{2}, f_{2+s}, f_{2+2s}, ...\}, \\
...  \\
v_{s,s} = \mbox{mean-pooling} \{f_{s}, f_{2s}, f_{3s}, ...\},
\end{array} 
\end{equation}
Together with the feature $\overline{v}$ obtained by mean pooling over the full sequence, skip sampling with a stride of $s$ produces $s+1$ training instances for the subsequent video-level feature augmentation.

\textbf{Video-level feature augmentation.}
Adding tiny perturbations to image pixels are imperceptible to humans. In a similar spirit, we want our video relevance predication system to ignore minor perturbations unconsciously introduced during feature extraction. 
To that end, we further employ perturbation-based data augmentation over each video-level feature generated from frame-level feature augmentation.
Given a $d$-dimensional video-level feature $v \in \mathbb{R}^d$, tiny Gaussian noises are randomly generated and selectively injected into the individual elements of the vector. More precisely, the perturbed feature $v^{*}$ is generated by: 
\begin{equation} \label{eq:video-da}
\begin{array}{l}
m \sim Bernoulli(p), \\
e \sim N_d(\mu, \sigma^2 I_{d}), \\
v^{*} = v + \epsilon \cdot m \circ e,
\end{array} 
\end{equation}
where $m$, as a mask, is a vector of independent Bernoulli random variables each of which has probability $p=0.5$ of being 1, which controls how many elements in the video-level feature are perturbed. The variable $e$ is a noise vector sampled from a multivariate Gaussian, parameterized by mean $\mu$ and covariance matrix $\sigma^2 I_{d}$, where $I_{d}$ is a $d\times d$ identity matrix. The mean and the standard deviation are estimated from the dataset. We empirically use $\epsilon=1$ to control the noise intensity. The symbol $\circ$ indicates element-wise multiplication.

After applying multi-level feature augmentation over a specific video $v$ having $n$ frame-level feature vectors, we obtain $s+1$ training instances $\{\overline{v}^*, v_{s,1}^*, v_{s,2}^*, ..., v_{s,s}^* \}$ for the supervised learning.
Notice that frame-level and video-level feature augmentation can be independently used to augment training data.
For instance, we can only perform video-level feature augmentation to generate a new sample for training if the frame-level feature of a video is unavailable.

Once our ranking-oriented feature re-learning model with the feature-level augmentation is trained on the ground-truth data with respect to a specific task, each video can be represented in a new video feature space that better reflects task-specific video relevance.

\subsection{Video Relevance Prediction}

\subsubsection{Two strategies}

In the context of the Hulu Content-based Video Relevance Prediction Challenge, the key is to predict relevance between a seed video and a candidate video.
Depending on whether a candidate video is known to be relevant to another candidate video, we consider two scenarios. In the first scenario, the relationship of a candidate video to another candidate video is unknown. Consequently, one has to fully count on the provided video features to predict the relevance between a given seed video and a candidate video. While in the second scenario, some candidate videos are known to be relevant with respect to some other candidate videos. Such a relationship might be exploited to improve video relevance prediction. Accordingly, different strategies are applied in different scenarios.

\textbf{Strategy 1}. 
In this strategy, we rely exclusively on the re-learned features.
In particular, given a seed video $v_s$ and a candidate video $v_c$, we estimate their video relevance by the cosine similarity in the re-learned video space:
\begin{equation}\label{eq:fresh_rel}
r(v_s, v_c) = cs_\phi(v_s, v_c).
\end{equation}

\textbf{Strategy 2}. Suppose the relationship of a candidate video to another candidate video is known, we exploit this extra clue to improve video relevance prediction. We hypothesize that if relevant videos of a candidate video are relevant to a given seed video, the candidate video is also likely to be relevant to the seed video. We implement our hypothesis by extending Eq. \ref{eq:fresh_rel} to include the relevance of the seed video to the top $n$ relevant videos of a candidate video $v_c$ as follows: 
\begin{equation}\label{eq:established_rel}
r(v_s,v_c) = cs_\phi(v_s,v_c) + \sum_{i=1}^{n} cs_\phi(v_s, v_{c,r}^{i}),
\end{equation}
where $v_{c,r}^{i}$ indicates the $i$-th relevant video with respect to the candidate video $v_c$.

Given a set of candidate videos, denoted by $\mathcal{V}$, we sort the candidate videos in descending order according to their relevance with respect to a given seed video. More formally, we solve the following optimization problem,
\begin{equation}\label{eq:known_rel}
\underset{v_c \in \mathcal{V}}{\max}~r(v_s, v_c),
\end{equation}
and consequently recommend the top $k$ videos.

\subsubsection{Time Complexity Analysis}

Once the model is trained, new features of each video in the candidate set $\mathcal{V}$ can be precomputed. Hence, our time complexity analysis focuses on the computation with respect to a seed video given on the fly. The computation consists of two parts, \ie feature projection for the seed video and relevance computation per candidate video in $\mathcal{V}$. The time complexity of feature projection is $O(d \times p )$, where $d$ and $p$ indicates the dimensionality of the original visual feature and projected feature, respectively. The time complexity of relevance computation depends on the relevance prediction strategy. It is $O(p)$ for the first strategy, and $O(p \times n)$ for the second strategy. Note that typically we have $n \ll |\mathcal{V}|$. Hence, strategy 1 has a linear complexity with respect to dataset size while the complexity for strategy 2 is sublinear.

%% file: table-notations.tex
\begin{table} [tb!]
\renewcommand{\arraystretch}{1.2}
\caption{Main notations defined in this paper.
}
\label{tab:notation}
\centering
 \scalebox{0.95}{
     \begin{tabular}{@{} l*{2}{l} @{}}
\toprule
Notation    && Description \\
\midrule  
    $v$                     &&   A video and its original video feature vector.   \\
    $d$                     &&   The dimensionality of an original video feature.  \\
    $cs(,)$                 &&   Cosine similarity in terms of the original video feature.  \\
    $\phi(v)$               &&   A re-learned video feature of video $v$.  \\
    $p$                     &&   The dimensionality of a re-learned video feature.  \\
    $cs_{\phi}(,)$          &&   Cosine similarity in terms of the re-learned video feature.  \\
    $v^+$                   &&   A video which is relevant with video $v$.  \\
    $v^-$                   &&   A video which is irrelevant with video $v$.  \\
    $\mathcal{D}$           &&   A set of relevant video pairs.  \\
    $\mathcal{T}$           &&   A set of triplets generated from $\mathcal{D}$.  \\
    $f_i$                   &&   A feature vector of the $i$-th video frame. \\
    $v_{s,i}$               &&   The $i$-th novel video generated by frame-level feature \\
                            &&   augmentation with stride of $s$ for the given video $v$. \\
    $v^*$                   &&   A video generated by video-level feature augmentation \\
                            &&   for the given video $v$. \\
    $\mathcal{V}$           &&   A set of candidate videos.  \\
\bottomrule
\end{tabular}
 }
\end{table}

%% file: eval.tex
\subsection{Experimental Setup} 

\textbf{Hulu challenge datasets.}
In order to verify the viability of the proposed feature re-learning solution, we use the TV-shows and Movies datasets provided by HULU in the context of the Content-based Video Relevance Prediction Challenge\footnote{\url{https://github.com/cbvrp-acmmm-2018/cbvrp-acmmm-2018}}.
Each dataset has been divided into three disjoint subsets for training, validation and test. 
Detailed data split is as follows: training / validation / test of 3,000 / 864 / 3,000 videos for the TV-shows dataset and 4,500 / 1,188 / 4,500 videos for the Movies dataset.
All videos are TV-show or movie trailers instead of full-length videos.
%

For each video in the training and validation set, it is associated with a list of relevant videos as ground truth derived from implicit viewer feedbacks. 
The relevant video list of a specific video $v$ is denoted as $ R_v = [v_r^1, v_r^2, ..., v_r^m]$, where $v^r_i$ indicates the video ranked at the $i$-th position in $R_v$ and $m$ is the number of the relevant videos. 
Notice that the ground truth of the test set is non-public. We have to submit our results to the task organizers and get performance scores back. It is thus impractical to evaluate every detail of the proposed model on the test set. We conduct most of the experiments with performance scores calculated on the validation set, unless otherwise stated. 

Concerning the set of candidate videos to be recommended, we follow the previous works \cite{hulu-baseline,ourmm18}, using the union of train and validation videos when evaluating on the validation set, and the union of train, validation and test videos when evaluating on the test set.


The video-wise relationship used in strategy 2 might be unavailable in practice, \eg in the cold-start scenario. So we use strategy 1 as the default choice for predicting video relevance unless otherwise stated.

As aforementioned, the HULU challenge does not provide original videos. Instead, two pre-computed features, \ie frame-level features and video-level features, are provided. 
Specifically, for frame-level features, videos are first decoded at 1 fps. Then decoded frames are fed into the InceptionV3 networks \cite{abu2016youtube} trained on ImageNet dataset \cite{ILSVRCarxiv14}, and the ReLU activations with 2,048 dimensions of the last hidden layer are used as the frame-level feature. 
For video-level features, the C3D model \cite{tran2015learning} trained on Sports1M dataset \cite{karpathy2014large} are leveraged. Each video is decoded at 8 fps and the activations of pool5 layer with 512 dimensions are utilized the final video clip feature.
For the ease of reference, we term the two features as Inception-v3 and C3D respectively.

\input{table-relearn}

\input{table-structure}

\textbf{Performance metrics.}
Following the evaluation protocol of the Content-based Video Relevance Prediction Challenge\cite{hulu-baseline}, we report two rank-based performance metrics, \ie recall@k (k = 50, 100, 200, 300) and hit@k (k = 5, 10, 20, 30). 
The performance for a specific seed video $v$ is computed as follows:
\begin{equation}
recall@k = \frac{\left | R_v \cap \widetilde{R_{vk}} \right |}{\left | R_v \right |},
\end{equation}
\begin{equation}
hit@k =\left\{\begin{matrix}
1, & if \ recall@k > 0
\\ 
0, & otherwise
\end{matrix}\right.,
\end{equation}
where $\widetilde{R_{vk}}$ denotes the top $k$ recommended videos from the candidate video set.
In practice, a user tends to browse top ranked videos in the first few pages, so hit with smaller $k$ better reflects a model's effectiveness. The overall performance is measured by summing up recall / hit scores over all the test videos.

\textbf{Implementations.}
PyTorch (\url{http://pytorch.org}) is used as our deep learning environment to implement the model. 
For the loss function, we empirically set the margin $m_1$ and $m_2$ in Eq. \ref{eq:ner_loss} as 0.2 and 0.05 respectively, and set $\alpha$ to be 1.
We train our model by stochastic gradient descent with Adam \cite{kingma2014adam}, and empirically set the initial learning rate to be 0.001 and batch size to be 32. 
We adopt a learning schedule as described in \cite{dong2018predicting}.
Once the validation loss does not decrease in three consecutive epochs, we divide the learning rate by 2. The early stop occurs if the validation performance does not improve in ten consecutive epochs. The maximal number of epochs is 50. 

\subsection{Experiment 1: Feature Re-learning} 

\begin{figure}[tb!]
\centering
 \subfigure [TV-shows] {
\noindent\includegraphics[width=0.47\columnwidth]{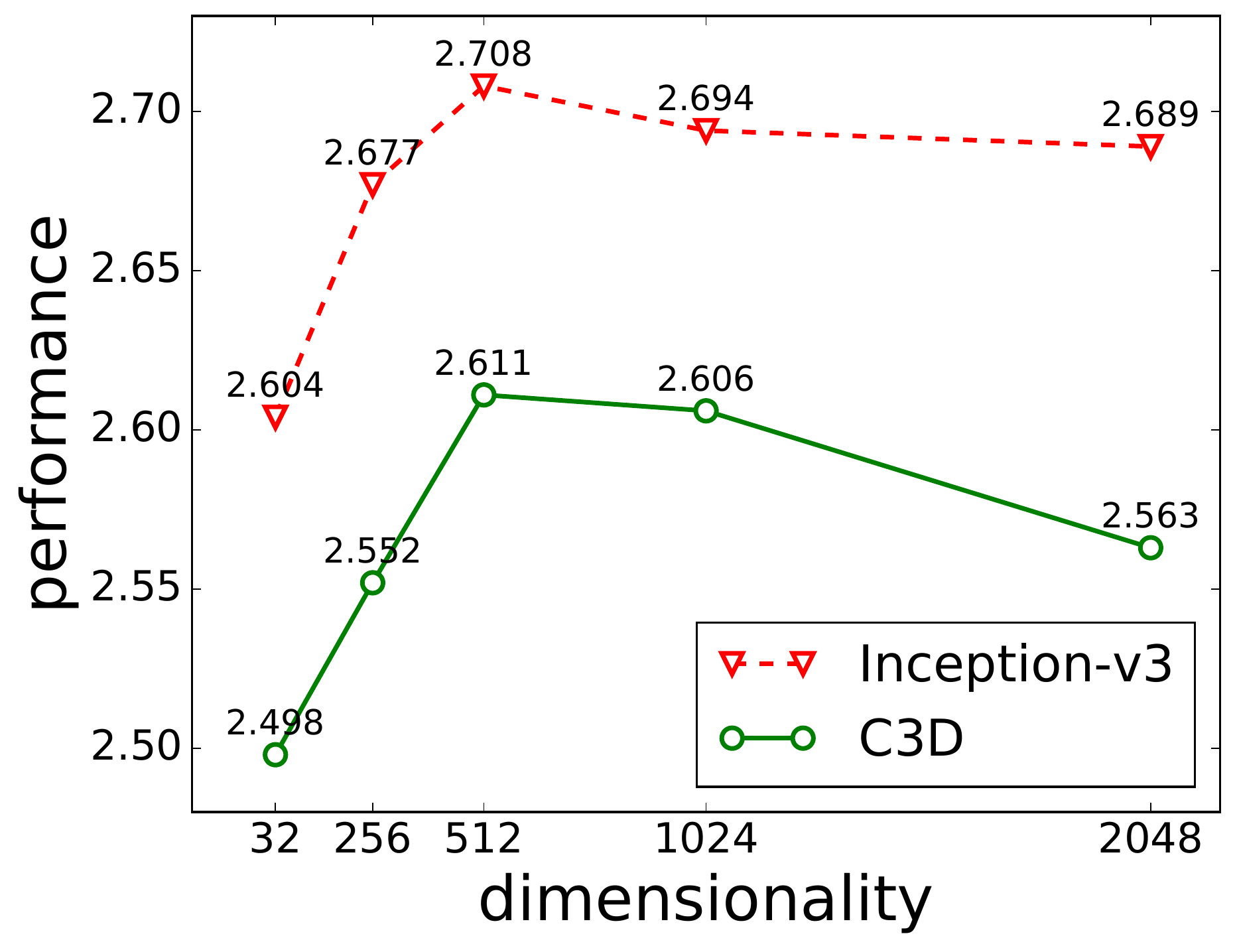}
\label{fig:frame_ad_shows}}
 \subfigure[Movies] {
\noindent\includegraphics[width=0.47\columnwidth]{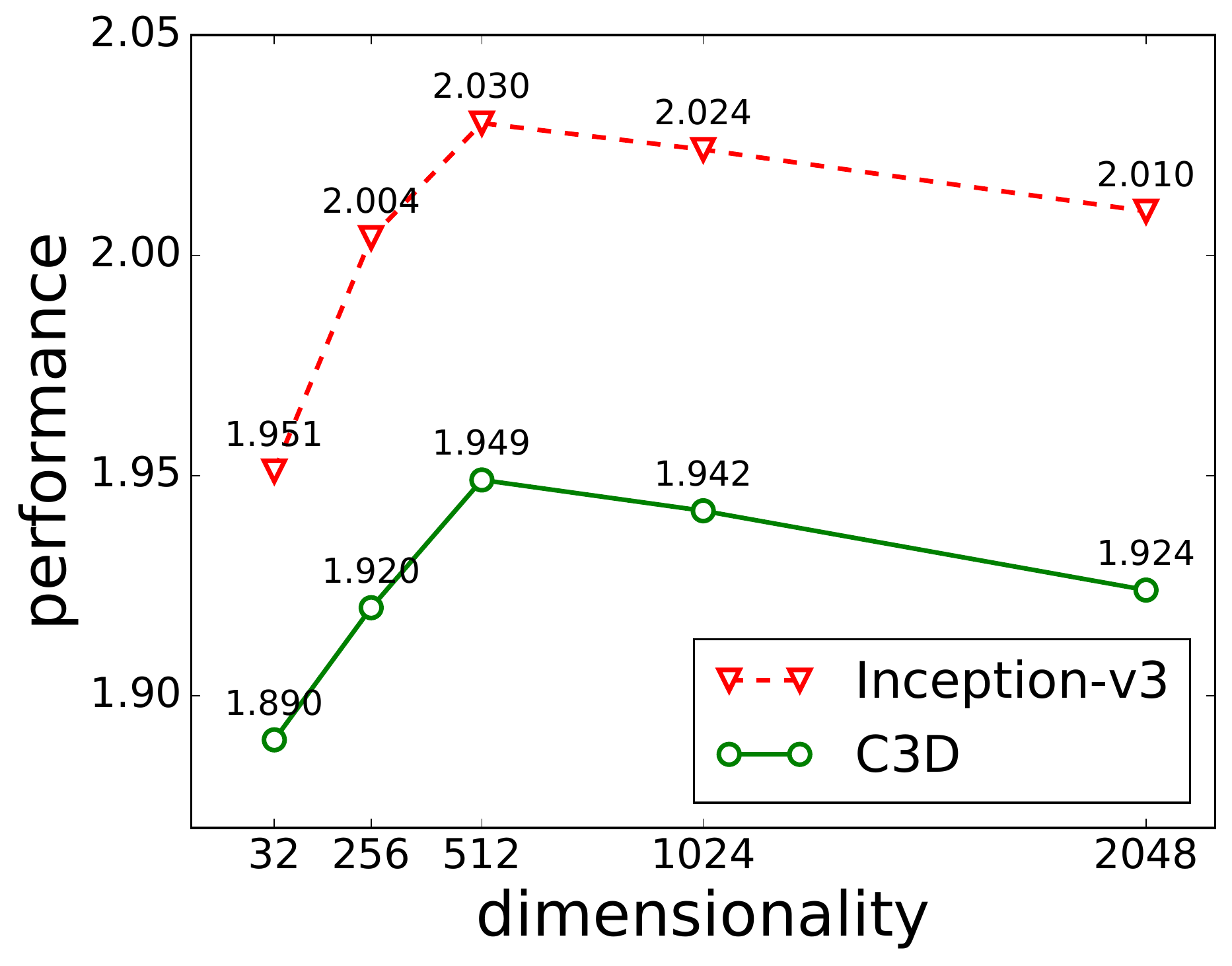}
\label{fig:frame_ad_movies}}
\caption{Performance curves of feature re-learning with varied dimensionality of the new video feature space on the (a) TV-shows and (b) Movies datasets. No data augmentation is used.}
 \label{fig:perf-relearn}
\end{figure}

In this experiment, we exploit the effectiveness of feature re-learning for video relevance prediction.
We study the relationship between the whole performance and the dimensionality of the re-learned feature space.
Specifically, for both Inception-v3 and C3D features, we compare the results of the dimensionality in the range of 32, 258, 512, 1024 and 2048 on the both TV-shows and Movies datasets. Note that the proposed data augmentation method is not employed here.
The results are shown in Fig \ref{fig:perf-relearn}. For the same dimensionality, the model using the Inception-v3 feature consistently outperforms the counterpart with the C3D feature. The best overall performance is reached with the dimensionality of 512, while the too small or too large dimensionality degrades the performance.
So we set the dimensionality of the re-learned feature space as 512 in the rest of the experiments.

Table \ref{tab:perf-re-learning} shows the performance of models with or without feature re-learning. The model without feature re-learning means directly utilizing the off-the-shelf feature to measure the video relevance.
On both datasets, re-learning consistently brings in a substantial performance gain.
Moreover, the advantage of feature re-learning model is feature independent. To be specific, the model with feature re-learning consistently outperforms its counterpart without feature re-learning for both given Inception-v3 and C3D features.
These results show the importance of feature re-learning for the video relevance prediction.

\input{table-loss}

Table \ref{tab:perf-structure} shows the performance of different architectures for feature projection. The one-layer architecture, as we have introduced in Eq. \ref{eq:affine-transform}, achieves the best overall performance with fewer parameters. Hence, we use it in the rest of our experiments.

For video relevance computation, we also tried the Euclidean distance instead of the cosine distance, but found the former less effective. Under the same setting, \ie TV-shows with Inception-v3, the model with the Euclidean distance obtains an overall performance of 2.235. By contrast, the model with the cosine distance has a higher score of 2.708.

\subsection{Experiment 2: Comparison of Loss Functions} 

In order to verify the viability of our proposed negative-enhanced triplet ranking loss (NETRL), we compare it with commonly used ranking loss functions in this experiment, \ie contrastive loss, standard triplet ranking loss (TRL) and improved triplet ranking loss (ITRL).
ITRL improves TRL via hard negative mining \cite{faghri2017vse}, by selecting the most similar yet irrelevant video as the negative instance instead of a randomly sampled instance. 
The performance comparison is summarized in Table \ref{tab:perf-loss}.
Recall that contrastive loss only considers the absolute similarity, while the TRL and ITRL consider only the relative similarity. Our proposed NETRL loss considers both absolute and relative similarities, showing the best performance on both datasets.
%
%

\begin{figure}[tb!]
\centering
 \subfigure[TV-shows]{
\noindent\includegraphics[width=0.47\columnwidth]{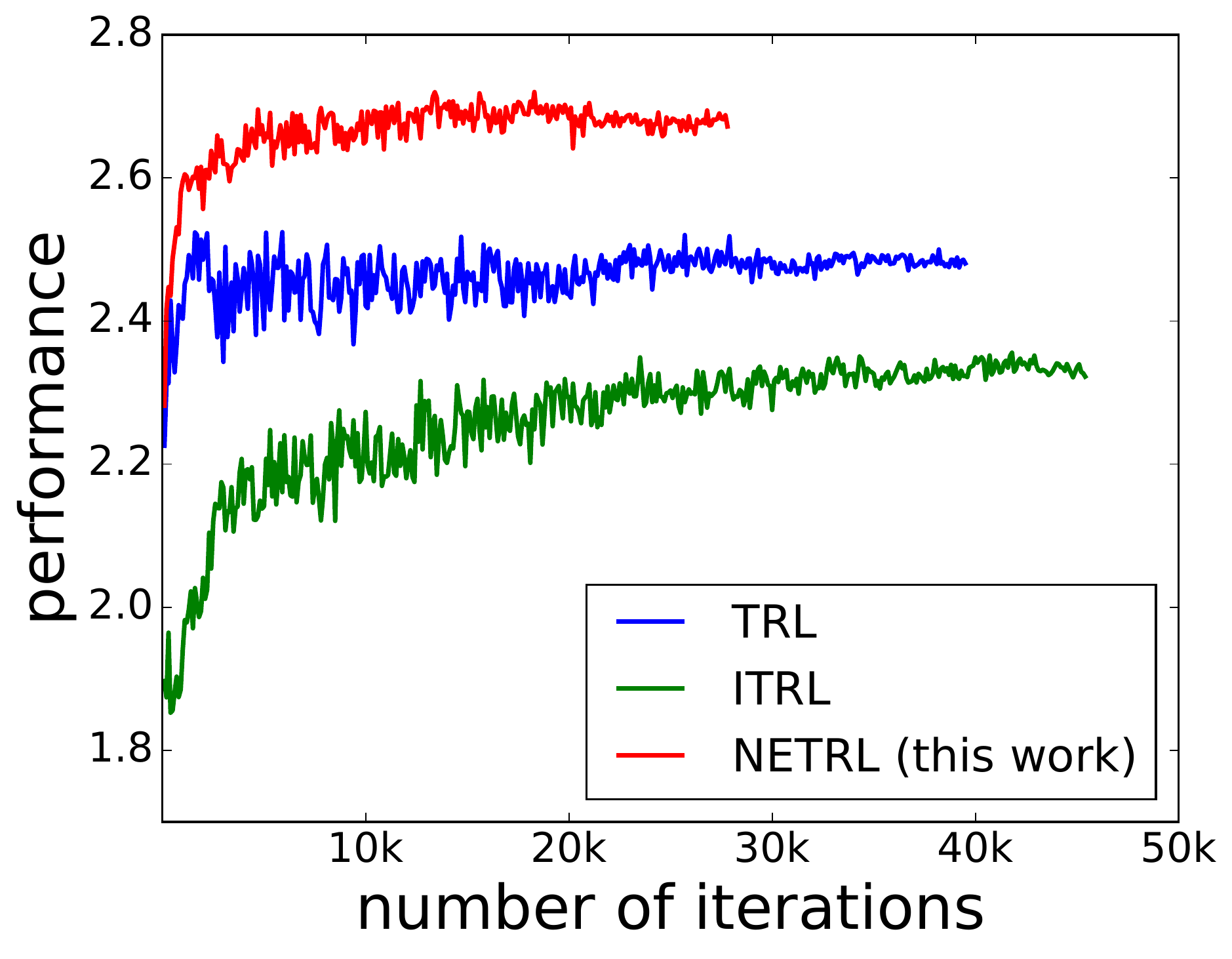}
\label{fig:perf-loss_shows}}
 \subfigure[Movies]{
\noindent\includegraphics[width=0.47\columnwidth]{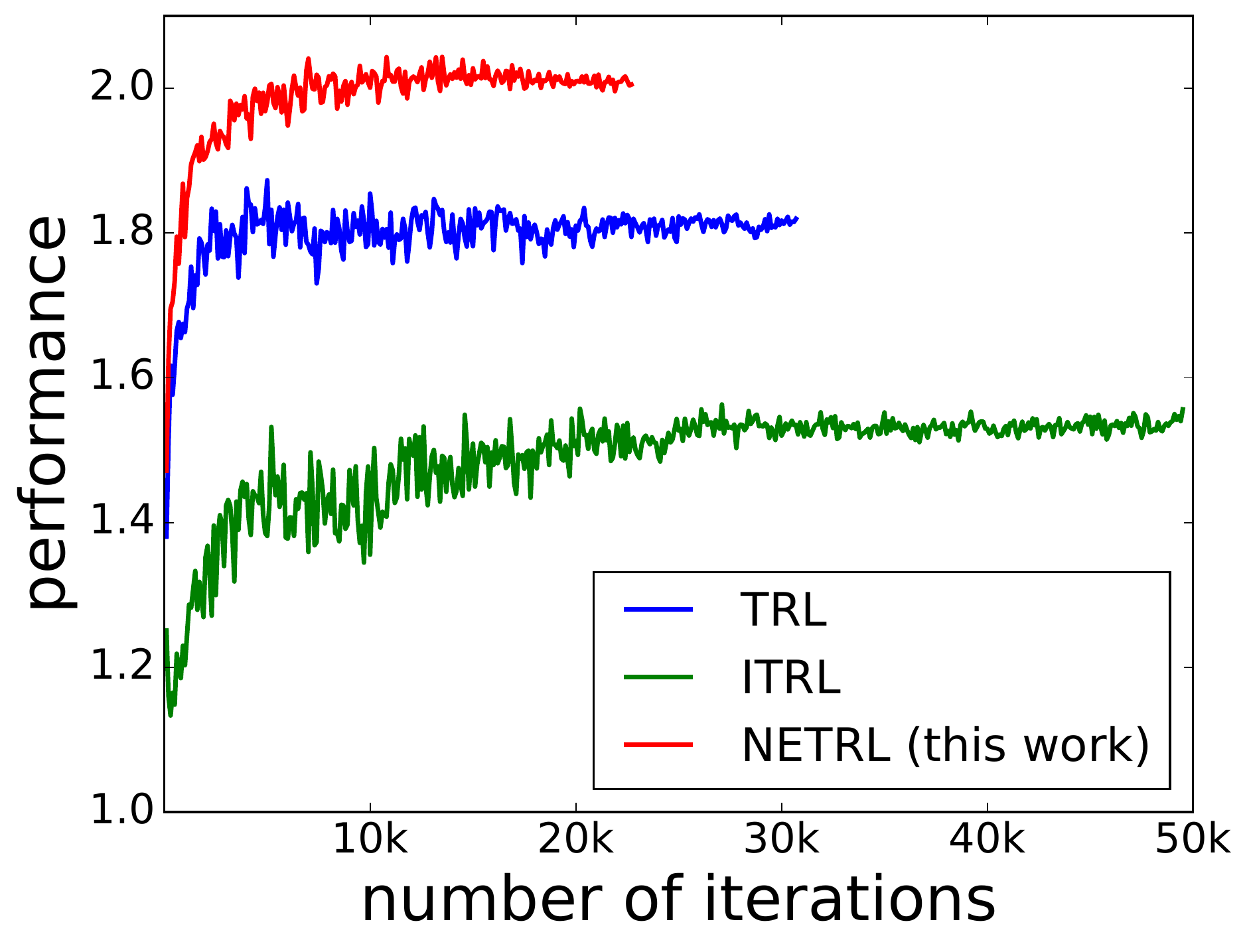}
\label{fig:perf-loss_movies}}
\caption{Performance curves of models trained with distinct losses on (a) TV-shows and (b) Movies datasets. Feature: Inception-v3. No data augmentation. Best viewed in color.}
 \label{fig:perf-loss}
\end{figure}

Interestingly, ITRL performs the worst, which is inconsistent with the existing results of other task \cite{cvpr2019-dual-dong}. We attribute the relatively lower performance of ITRL to the limited training data in our experiments. As noted in the original paper \cite{faghri2017vse}, ITRL requires more training iterations than TRL. So we compare the training behavior of TRL, ITRL and NETRL. 
Specifically, we record the validation performance every 100 iterations. As the performance curves in  Fig. \ref{fig:perf-loss_shows} and Fig. \ref{fig:perf-loss_movies} show, NETRL consistently outperforms the other losses. 
Moreover, it has the fastest convergence. 

Further, we test the stability of models trained with distinct losses, by reporting mean and standard deviation of the performance scores. As the performance changes drastically in the early stage of training, we take into account the performance scores after training over 5k iterations. As Table \ref{tab:perf_stability} shows, NETRL yields larger mean and lower variance, meaning the corresponding models are more stable.

\begin{table} [tb!]
\renewcommand{\arraystretch}{1.2}
\caption{The mean and standard deviation of the model performance with distinct ranking loss functions after training over 5k iterations. The larger mean and smaller standard deviation indicate better.
}
\label{tab:perf_stability}
\centering
 \scalebox{1.0}{
     \begin{tabular}{@{} lcc*{2}{c} @{}}
\toprule
\multirow{2}{*}{\textbf{Loss Function}}    && \multicolumn{2}{c}{\textbf{Dataset}} \\
\cmidrule(r){2-4}
    && TV-shows   & Movies \\
\midrule
    \textbf{TRL}            && 2.47$\pm$0.025 &  1.81$\pm$0.018  \\
    \textbf{ITRL}   && 2.29$\pm$0.053 &  1.51$\pm$0.042  \\
    \textbf{NETRL}  && \textbf{2.68$\pm$0.017} &  \textbf{2.01$\pm$0.014}  \\
\bottomrule
\end{tabular}
 }
\end{table}

\input{table-dataaug}

\begin{figure}[tb!]
\centering
 \subfigure[Effect of $m_1$ and $\alpha$ ($m_2=0.05$)]{
\noindent\includegraphics[width=0.47\columnwidth]{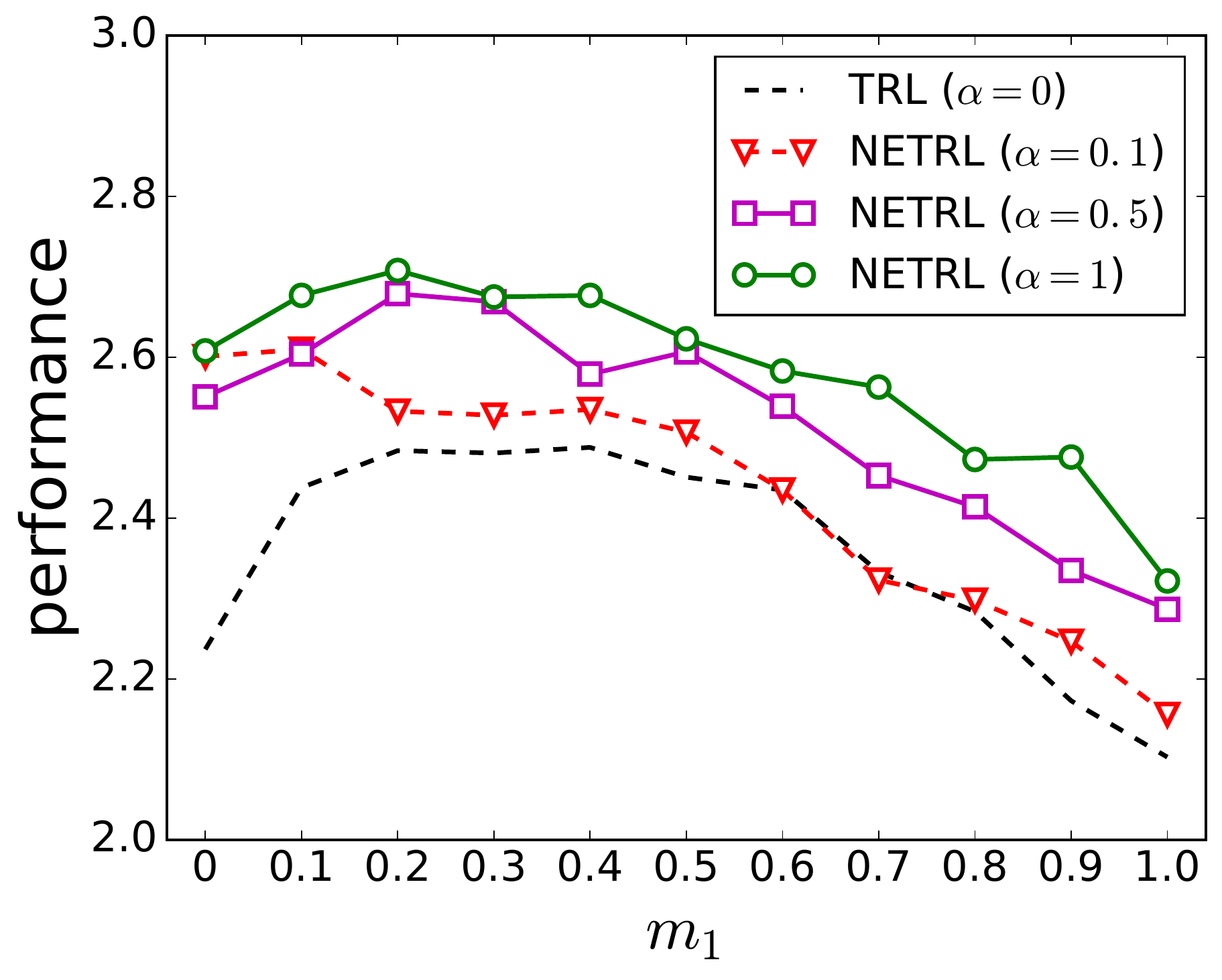}
\label{fig:hyperp-m1}}
 \subfigure[Effect of $m_2$ and $\alpha$ ($m_1=0.2$)]{
\noindent\includegraphics[width=0.47\columnwidth]{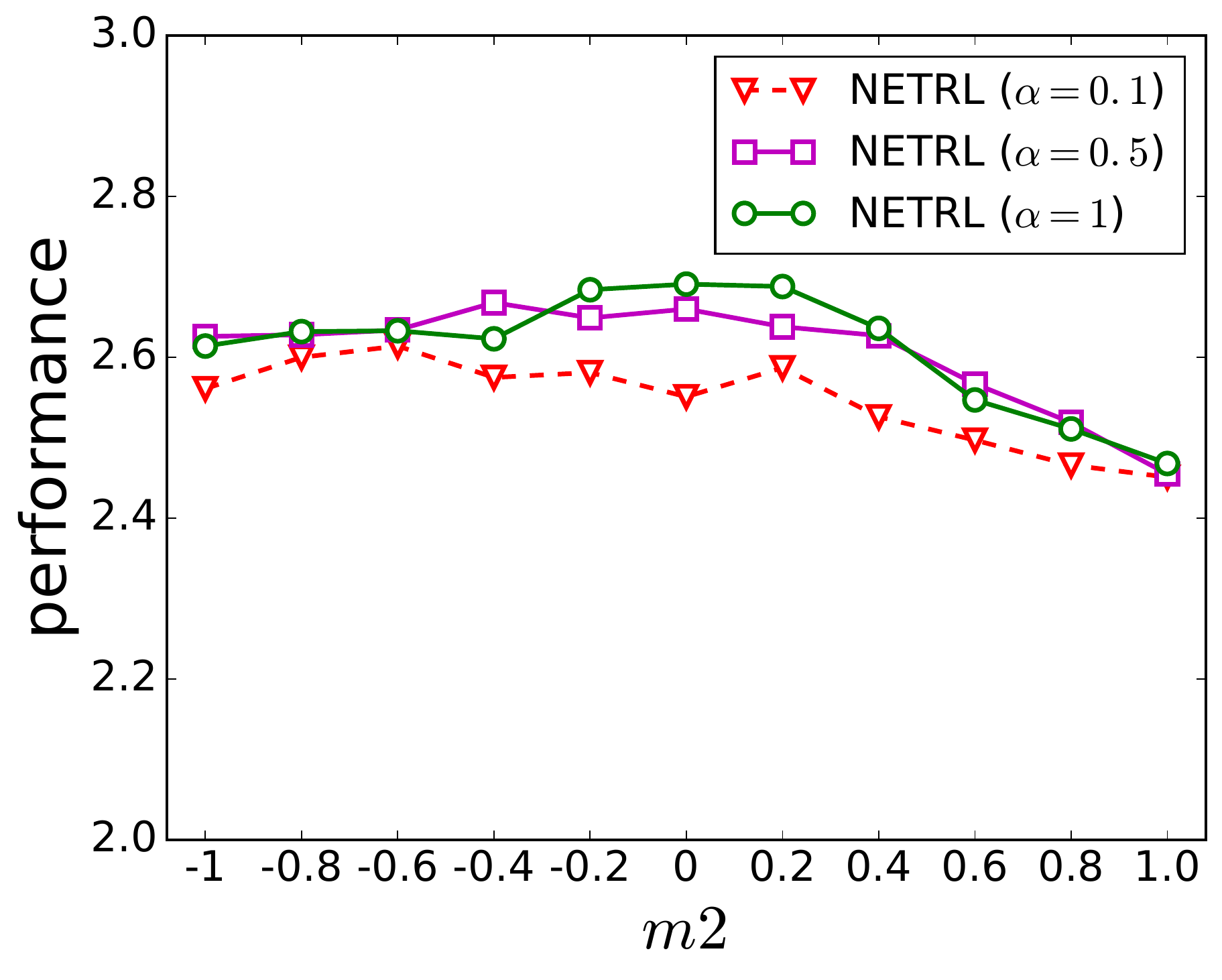}
\label{fig:hyperp-m2}}
\caption{The effect of the three hyper parameters, \ie $m_1$, $\alpha$ and $m_2$, in the proposed NETRL loss. Dataset: TV-shows. Feature: Inception-v3. No data augmentation. With the negative enhanced term omitted by setting $\alpha$ to $0$ or setting $m_2$ to $1$, NETRL is reduced to the classical TRL loss.}
 \label{fig:hyperp}
\end{figure}

Recall that NETRL is a weighted combination of two terms with three hyper parameters, \ie $m_1$, $\alpha$ and $m_2$. To figure out what actually works, we investigate the influence of these parameters. As shown in Fig. \ref{fig:hyperp-m1}, the curves corresponding to NETRL with varied $\alpha$ top the dashed curve corresponding to TRL, confirming the necessity of the negative enhanced term in NETRL. As for $m_1$, the curves go up first as $m_1$ increases, showing the positive effective of the margin. However, as the margin value becomes larger, it makes the learning process unnecessarily more difficult, and consequently makes the learned model less discriminative. While the optimal value of $m_1$ is clearly task-dependent, on our experimental data the peak performance is reached with $m_1=0.2$. As for $m_2$, since the maximum of the cosine similarity is $1$, the negative term is weakened as $m_2$ approaches $1$. Consequently, the performance curves go down, see Fig. \ref{fig:hyperp-m2}. Setting $m_2$ around 0 gives good performance. 
In addition, we tried to replace the second term with a similar constraint but on positive pairs, \ie $max(0, cs_\phi(v,v^+)-m_2)$, but found the performance worse. Under the same setting, \eg TV-shows with Inception-v3, its overall performance is 2.492, where the NETRL gives a higher performance of 2.708.
These results suggest that considering both relative and absolute similarity among videos pairs is important for training a video relevance prediction model.
%

\subsection{Experiment 3: Feature Augmentation}

\begin{figure}[tb!]
\centering
 \subfigure [TV-shows] {
\noindent\includegraphics[width=0.47\columnwidth]{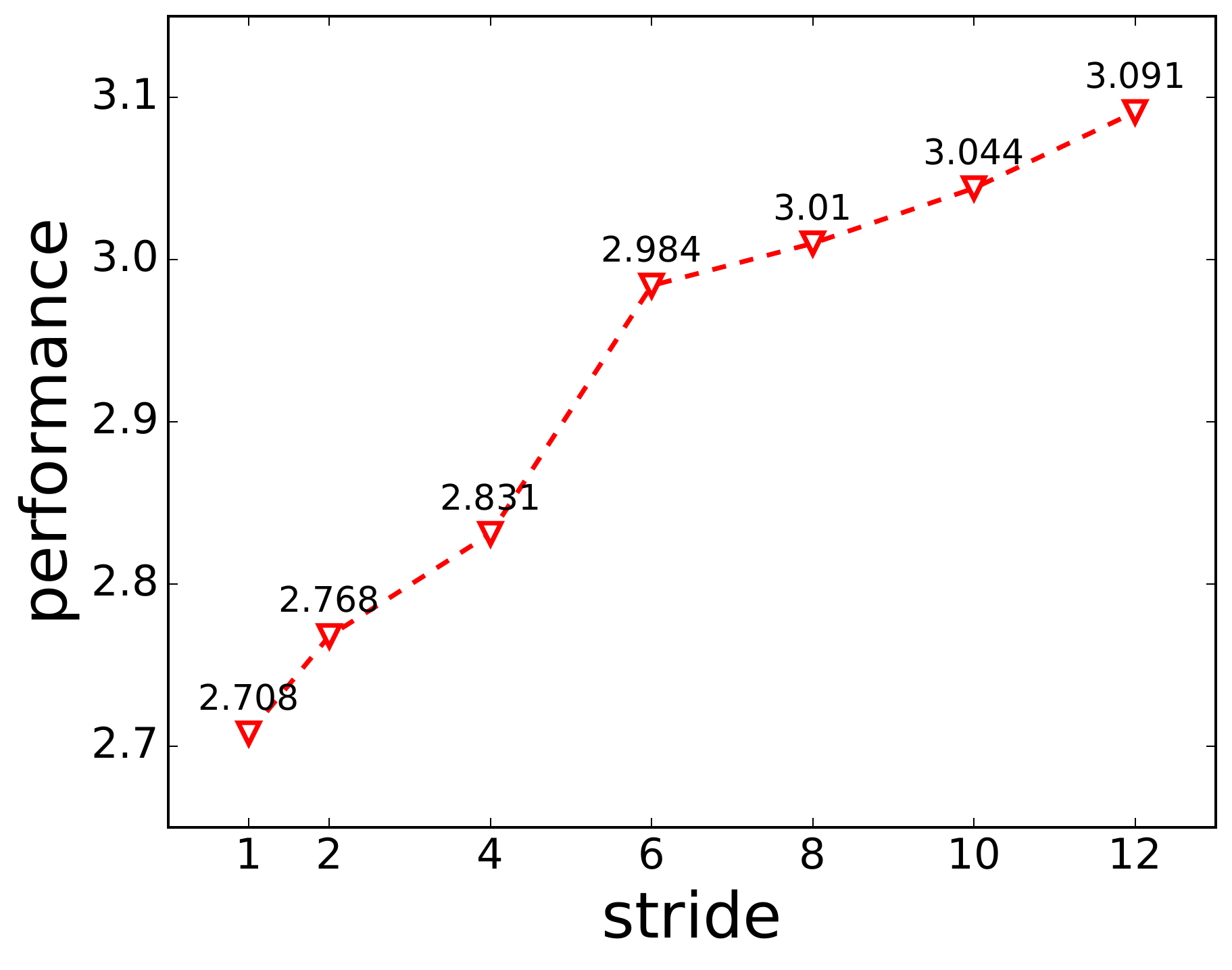}
\label{fig:frame_ad_shows}}
 \subfigure[Movies] {
\noindent\includegraphics[width=0.47\columnwidth]{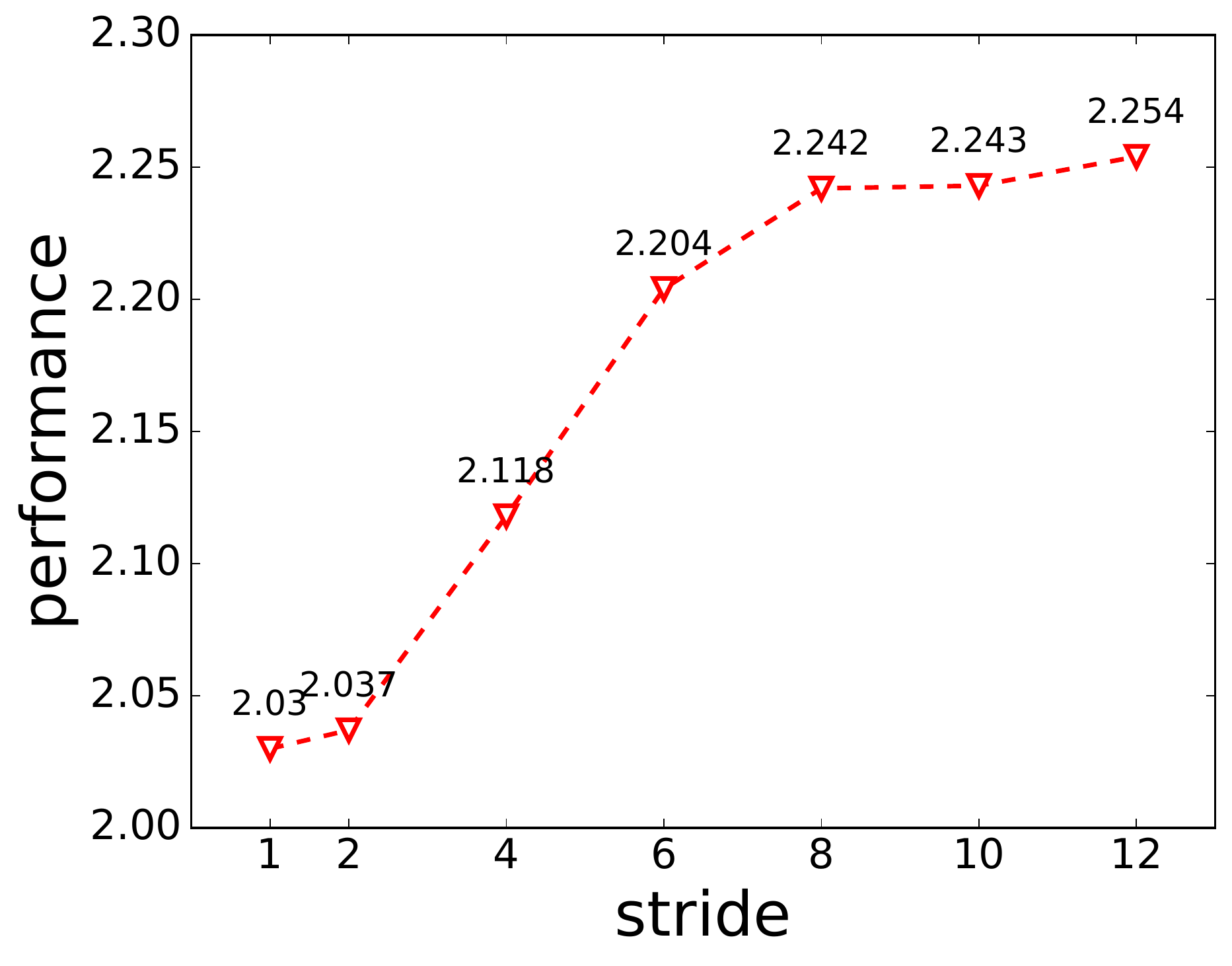}
\label{fig:frame_ad_movies}}
\caption{Performance curves of feature re-learning with data augmentation performed on the Inception-v3 feature. The starting point, $s=1$, means no data augmentation.}
 \label{fig:perf-frame-da}
\end{figure}

As the proposed multi-level augmentation strategy consists of frame-level and video-level feature augmentations, we first evaluate them individually and then evaluate the full augmentation method. 

Figure \ref{fig:perf-frame-da} shows the performance curves of feature re-learning model with frame-level feature augmentation as the stride increases on the TV-shows and Movies datasets. 
The rising curves justify the effectiveness of data augmentation for the frame-level feature.
The performance improvement is significant at the beginning while slows down when the stride is larger than 8.
The overall best performance is reached at $stride=12$. 
The detailed performances are shown in the first two rows for Inception-v3 in Table \ref{tab:perf-da}, where our model with the frame-level feature augmentation consistently outperforms the counterpart without any augmentation strategies in terms of all the evaluation metric. The result confirms the effectiveness of the frame-level feature augmentation.
Besides, we also exploit multiple skip samplings with the varied stride together, and we found some improvement gain.
With the multiple skip samplings with the stride of 8, 10 and 12 together, the whole performance on TV-shows and Movies are 3.092 and 2.289 respectively. However, the augmentation with multiple skip samplings makes the training time longer. For the balance of the performance and training time, we set the $stride=12$ as the default parameter in the data augmentation for frame-level features unless otherwise stated.

As for the video-level feature augmentation, we conduct the experiments with both Inception-v3 and C3D features.
For the frame-level feature, \ie Inception-v3, mean pooling is firstly conducted over the features to obtain the video-level feature before the augmentation.
As Table \ref{tab:perf-da} shows, our model with Inception-v3 obtains the whole performance of 2.802 and 2.109 on the TV-shows and Movies datasets, while the scores of its baseline without data augmentation are 2.708 and 2.030, respectively. Similar phenomenons are observed using the C3D features.
The results verify that the video-level feature augmentation is also meaningful for feature re-learning in the context of video recommendation.

Comparing the frame-level and video-level augmentations over the Inception-v3 feature (Row 2 and 3 in Table \ref{tab:perf-da}), we find the frame-level method gives a higher performance boost over the counterpart without data augmentation. It provides relative improvement of 14.1\% and 11.0\% on TV-shows and Movies, respectively, while the corresponding numbers of the video-level method are 3.5\% and 3.9\%. 

\input{table-video-da}

Concerning the effect of $\mu$ and $\sigma$ in Eq. \ref{eq:video-da} for video-level data augmentation, we choose $\sigma$ from $\{0.1, 1, 10\}$ with $\mu$ fixed to be $0$. As Table \ref{tab:perf-video-da} shows, using the parameters estimated from the training data performs the best.

An interesting phenomenon is that while using either frame-level or video-level augmentation alone is helpful, their combination (Row 4 in Table \ref{tab:perf-da}) degenerates the performance, reducing the overall score from 3.091 to 2.944. Our explanation for this phenomenon is that by skip sampling, the frame-level method produces new training samples that are relatively close to the decision boundary. While these samples help learning a more precise model for video relevance prediction, they are more sensitive to extra noise. With noise injected by the video-level method, the samples might incorrectly shift to the wrong side, and consequently affect the model.
Next, we study how the individual data augmentation methods and their combination behave when extra noises are artificially introduced to the test data.

\subsection{Experiment 4: Robust Analysis} \label{subsec:rebust}

\begin{figure}[tb!]
\centering
 \subfigure [TV-shows] {
\noindent\includegraphics[width=0.47\columnwidth]{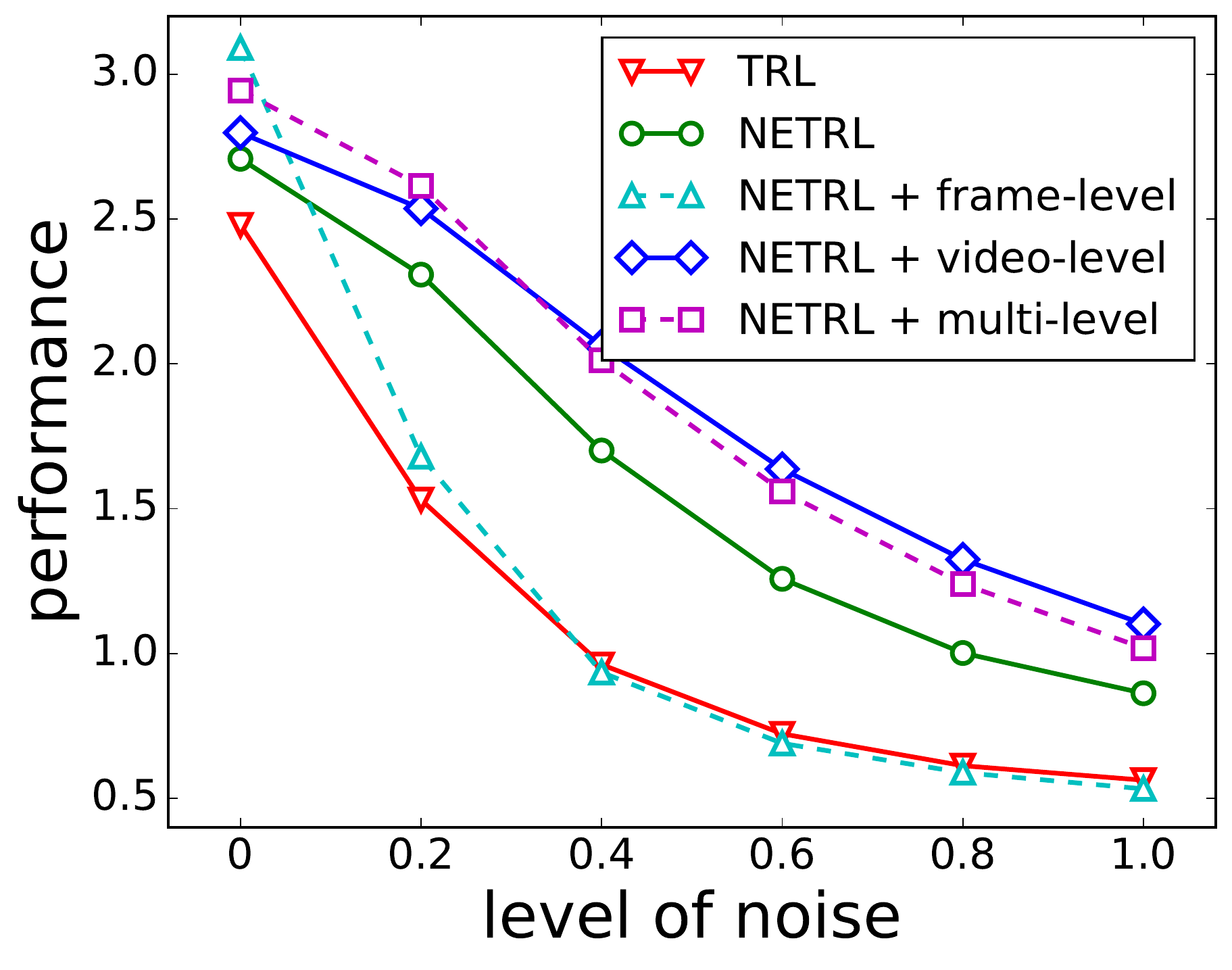}
\label{fig:robust_shows}}
 \subfigure[Movies] {
\noindent\includegraphics[width=0.47\columnwidth]{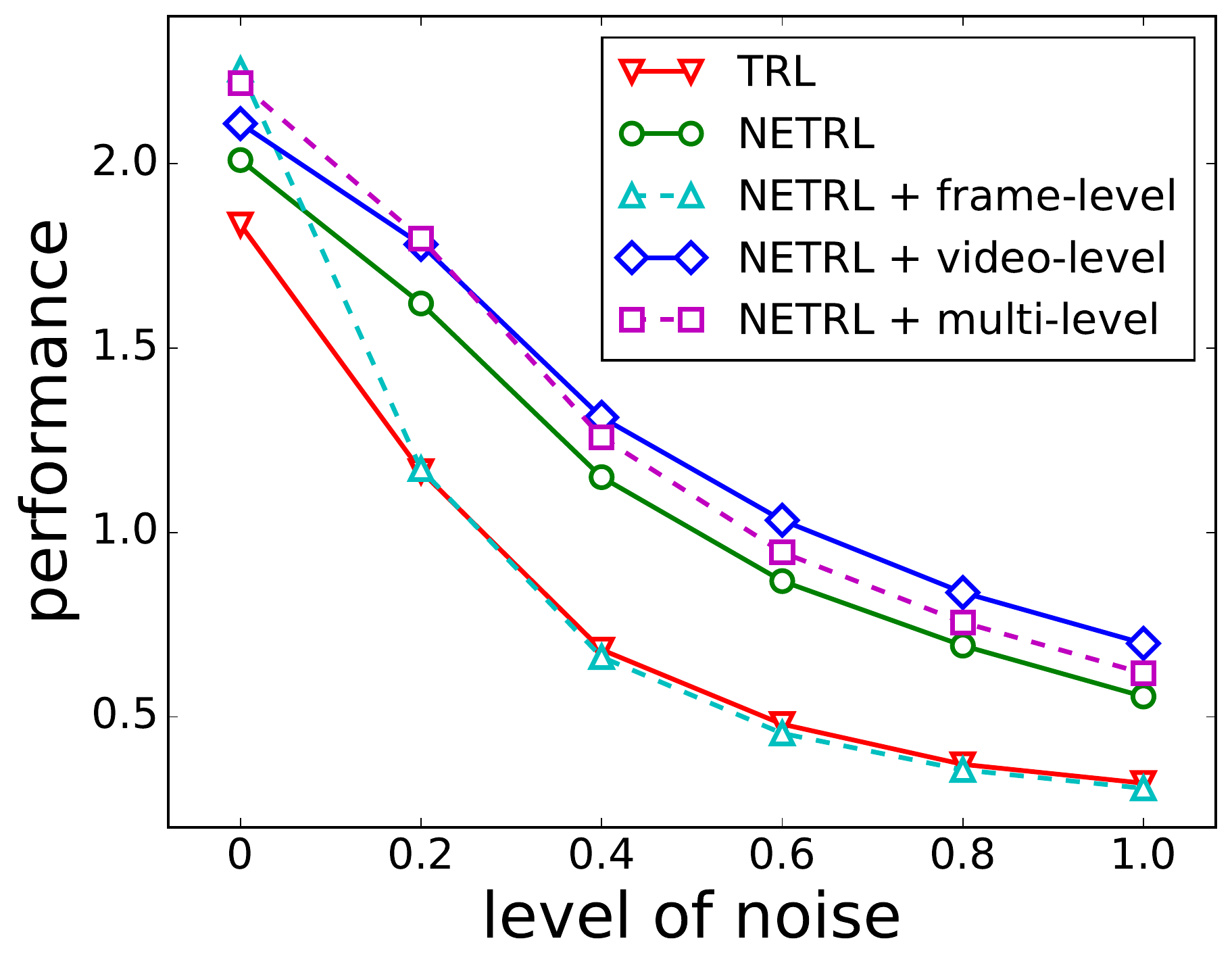}
\label{fig:robust_movies}}
\caption{Performance curves of different models with respect to artificially injected noise. For a given level of noise, the noise is multiplied with the corresponding coefficient. Feature: Inception-v3. The model trained with proposed multi-level feature augmentation best balances the performance and robustness.}
 \label{fig:robust}
\end{figure}

We now analyze the robustness of our method by adding extra noise to video-level features. Concretely, for each video, we add random noise sampled from a Gaussian distribution ($\mu=0$, $\sigma=1$).
The performance curves with respect to the level of noise are shown in Fig. \ref{fig:robust}. Unsurprisingly, for all methods, the performance goes down.
Comparing TRL and NETRL without any augmentation strategies, the curve of TRL drops more sharply, which shows the better robustness of our proposed NETRL loss function for video relevance prediction.
Considering NETRL with various augmentation strategies, the curve of frame-level feature augmentation drops fastest. Although our model with the frame-level augmentation performs well in the absence of extra noises, its robustness is the worst. The reverse result is observed for the video-level feature augmentation. It performs worse without adding any extra noise, while shows good robustness. 
Among them the multi-level feature augmentation best balances the model performance and robustness.

\input{sota}

%% file: table-relearn.tex
\begin{table*} [tb!]
\renewcommand{\arraystretch}{1.2}
\caption{Effectiveness of feature re-learning. For both Inception-v3 and C3D features, re-learning brings in substantial performance gain. }
\label{tab:perf-re-learning}
\centering
 \scalebox{1.0}{
     \begin{tabular}{@{} lcc*{10}{c} @{}}
\toprule
\multirow{2}{*}{\textbf{Dataset}}   &  \multirow{2}{*}{\textbf{Feature}} &  \multirow{2}{*}{\textbf{Re-Learning}} & \multicolumn{4}{c}{\textbf{hit@k}} &  & \multicolumn{4}{c}{\textbf{recall@k}} & \multirow{2}{*}{\textbf{Sum}} \\
         \cmidrule(r){4-7}  \cmidrule(r){9-12} 
    &   &   & k=5 & k=10 & k=20 & k=30 & &  k=50 & k=100 & k=200 & k=300 & \\
\midrule
    \multirow{4}{*}{\textbf{TV-shows}} &   \multirow{2}{*}{Inception-v3} 
          & \xmark   & 0.234 &  0.316 &  0.397 &  0.448     &&  0.083 & 0.124 & 0.192 & 0.244 & 2.038 \\ 
        & & \cmark     & \textbf{0.285} & \textbf{0.391} & \textbf{0.483} & \textbf{0.539} && \textbf{0.138} & \textbf{0.208} & \textbf{0.299} & \textbf{0.365} & \textbf{2.708} \\
        \cmidrule(r){2-13} 
        &  \multirow{2}{*}{ C3D}      
          & \xmark       & 0.234 & 0.313 & 0.409 & 0.488     &&  0.092 & 0.145 & 0.216 & 0.267 & 2.164 \\
        & & \cmark       & \textbf{0.269} & \textbf{0.362} & \textbf{0.468} & \textbf{0.544} && \textbf{0.130} & \textbf{0.198} & \textbf{0.288} & \textbf{0.352} & \textbf{2.611} \\
 \midrule
    \multirow{4}{*}{\textbf{Movies}} &     \multirow{2}{*}{Inception-v3} 
          & \xmark   & 0.141 & 0.185 & 0.248 & 0.291     &  &  0.072 & 0.099 &0.137 & 0.167 & 1.340 \\
        & & \cmark       & \textbf{0.187} & \textbf{0.254} & \textbf{0.341} & \textbf{0.407}  && \textbf{0.118} & \textbf{0.173} & \textbf{0.248} & \textbf{0.302} & \textbf{2.030} \\
        \cmidrule(r){2-13} 
        &  \multirow{2}{*}{ C3D}      
          & \xmark      & 0.140 & 0.193 & 0.271 & 0.316   &&  0.084 & 0.112 & 0.160 & 0.196 & 1.472 \\
        & & \cmark      & \textbf{0.180} & \textbf{0.251} & \textbf{0.332} & \textbf{0.392}   &&  \textbf{0.116} & \textbf{0.164} & \textbf{0.231} & \textbf{0.283} & \textbf{1.949} \\
\bottomrule
\end{tabular}
 }
\end{table*}

%% file: table-structure.tex
\begin{table*} [tb!]
\renewcommand{\arraystretch}{1.2}
\caption{Performance of different feature projection architectures. Feature:  Inception-v3. Numbers in the parenthesis denotes the sizes of input, hidden, and output layers in a fully connected  network (FCN).}
\label{tab:perf-structure}
\centering
 \scalebox{0.95}{
     \begin{tabular}{@{} llccc*{9}{c} @{}}
\toprule
\multirow{2}{*}{\textbf{Dataset}}  &   \multirow{2}{*}{\textbf{Feature projection architecture}}  & \multirow{2}{*}{\textbf{\#Parameters}} & \multicolumn{4}{c}{\textbf{hit@k}} &  & \multicolumn{4}{c}{\textbf{recall@k}} & \multirow{2}{*}{\textbf{Sum}} \\
         \cmidrule(r){4-8}  \cmidrule(r){9-12} 
    & &  &  k=5 & k=10 & k=20 & k=30 & &  k=50 & k=100 & k=200 & k=300 & \\
\midrule
    \multirow{3}{*}{\textbf{TV-shows}} &  
          One-layer FCN (2048-512)  & 1.05 millions  & \textbf{0.285} & \textbf{0.391} & \textbf{0.483} & \textbf{0.539} && 0.138 & 0.208 & 0.299 & 0.365 & \textbf{2.708} \\ 
        & Two-layer FCN (2048-512-512)  & 1.31 millions  & 0.262 & 0.360 & 0.450 & 0.506 && 0.137 & 0.215 & 0.319 & 0.393 & 2.642 \\
        & Two-layer residual FCN (2048-512-512) &  1.31 millions & 0.280 & 0.367 & 0.458 & 0.515 && \textbf{0.140} & \textbf{0.225} & \textbf{0.326} & \textbf{0.393} & 2.704 \\ 
 \midrule
    \multirow{3}{*}{\textbf{Movies}} & 
          One-layer FCN (2048-512) & 1.05 millions  & \textbf{0.187} & \textbf{0.254} & \textbf{0.341} & \textbf{0.407}  && \textbf{0.118} & 0.173 & 0.248 & 0.302 & \textbf{2.030} \\
        & Two-layer FCN (2048-512-512) &  1.31 millions  & 0.183 & 0.237 & 0.321 & 0.382 && 0.114 & 0.175 & \textbf{0.258} & \textbf{0.319} & 1.989 \\
        & Two-layer residual FCN (2048-512-512) &  1.31 millions & 0.185 & 0.252 & 0.337 & 0.392 && \textbf{0.118} & \textbf{0.177} & 0.253 & 0.314 & 2.028\\
\bottomrule
\end{tabular}
 }
\end{table*}

%% file: table-loss.tex
\begin{table*} [tb!]
\renewcommand{\arraystretch}{1.2}
\caption{Performance comparison of feature re-learning with different loss on the validation set. No data augmentation. Our proposed negative-enhanced triplet ranking loss performs the best.}
\label{tab:perf-loss}
\centering
 \scalebox{1.0}{
     \begin{tabular}{@{} lcc*{10}{c} @{}}
\toprule
\multirow{2}{*}{\textbf{Dataset}}   &  \multirow{2}{*}{\textbf{Feature}} &  \multirow{2}{*}{\textbf{Loss Function}} & \multicolumn{4}{c}{\textbf{hit@k}} &  & \multicolumn{4}{c}{\textbf{recall@k}} & \multirow{2}{*}{\textbf{Sum}} \\
         \cmidrule(r){4-7}  \cmidrule(r){9-12} 
    &   &   & k=5 & k=10 & k=20 & k=30 & &  k=50 & k=100 & k=200 & k=300 \\
\midrule
    \multirow{6}{*}{\textbf{TV-shows}} &   \multirow{3}{*}{Inception-v3} 
        & Contrastive loss   & 0.231 & 0.326 & 0.424 & 0.495  &&  0.118 & 0.185 & 0.282 & 0.351 & 2.412 \\ [1pt]
        & & TRL                 & 0.245 & 0.333 & 0.440 & 0.503     &  &  0.126 & 0.193 & 0.290 & 0.354 & 2.484 \\
        & & ITRL      & 0.240 & 0.310 & 0.411 & 0.478     &  &  0.120 & 0.180 & 0.260 & 0.316 & 2.315 \\
        & & NETRL     & \textbf{0.285} & \textbf{0.391} & \textbf{0.483} & \textbf{0.539} && \textbf{0.138} & \textbf{0.208} & \textbf{0.299} & \textbf{0.365} & \textbf{2.708}  \\
       
        \cmidrule(r){2-13} 
        &  \multirow{3}{*}{ C3D}      
        & Contrastive loss      & 0.225 & 0.317 & 0.438 & 0.505     &&  0.110 & 0.172 & 0.259 & 0.328  & 2.354 \\
        & & TRL                 & 0.247 & 0.325 & 0.433 & 0.512     &  &  0.119 & 0.183 & 0.275 & 0.343   & 2.437 \\
        & & ITRL      & 0.199 & 0.281 & 0.382 & 0.462     &  &  0.085 & 0.127 & 0.189 & 0.242   & 1.967  \\
        & & NETRL     & \textbf{0.269} & \textbf{0.362} & \textbf{0.468} & \textbf{0.544} && \textbf{0.130} & \textbf{0.198} & \textbf{0.288} & \textbf{0.352} & \textbf{2.611} \\

\midrule
    \multirow{6}{*}{\textbf{Movies}} &   \multirow{3}{*}{Inception-v3} 
        & Contrastive loss   & 0.146 & 0.196 & 0.274 & 0.332    &&  0.095 & 0.148 & 0.216 & 0.271 & 1.678\\ [1pt]
        & & TRL                 & 0.155 & 0.220 & 0.303 & 0.371     &  &  0.102 & 0.158 & 0.234 & 0.292    & 1.835 \\
        & & ITRL      & 0.141 & 0.194 & 0.284 & 0.338     &  &  0.088 & 0.126 & 0.178 & 0.218    & 1.567 \\
        & & NETRL     & \textbf{0.187} & \textbf{0.254} & \textbf{0.341} & \textbf{0.407} && \textbf{0.118} & \textbf{0.173} & \textbf{0.248} & \textbf{0.302} & \textbf{2.030} \\
       
        \cmidrule(r){2-13}
        &  \multirow{3}{*}{ C3D}      
        & Contrastive loss      & 0.131 & 0.191 & 0.273 & 0.343     &&  0.097 & 0.144 & 0.210 & 0.260 & 1.649 \\
        & & TRL                 & 0.166 & 0.224 & 0.324 & 0.383     &  &  0.106 & 0.153 & 0.219 & 0.278 & 1.853 \\
        & & ITRL      & 0.110 & 0.161 & 0.231 & 0.269     &  &  0.061 & 0.090 & 0.127 & 0.163 & 1.212 \\
        & & NETRL     & \textbf{0.180} & \textbf{0.251} & \textbf{0.332} & \textbf{0.392} && \textbf{0.116} & \textbf{0.164} & \textbf{0.231} & \textbf{0.283} & \textbf{1.949} \\

\bottomrule
\end{tabular}
 }
\end{table*}

%% file: table-dataaug.tex
\begin{table*} [tb!]
\renewcommand{\arraystretch}{1.2}
\caption{ Effectiveness of data augmentation. The model with data augmentation gives better performance.}
\label{tab:perf-da}
\centering
 \scalebox{1.0}{
     \begin{tabular}{@{} lccc*{10}{c} @{}}
\toprule
\multirow{2}{*}{\textbf{Dataset}}   &  \multirow{2}{*}{\textbf{Feature}} &  \multicolumn{2}{c}{\textbf{Data augmentation}} & \multicolumn{4}{c}{\textbf{hit@k}} &  & \multicolumn{4}{c}{\textbf{recall@k}} & \multirow{2}{*}{\textbf{Sum}} \\
    \cmidrule(r){3-4}     \cmidrule(r){5-8}  \cmidrule(r){10-13} 
    &   & frame-level & video-level   & k=5 & k=10 & k=20 & k=30 & &  k=50 & k=100 & k=200 & k=300 \\
\midrule
    \multirow{6}{*}{\textbf{TV-shows}} &   \multirow{4}{*}{Inception-v3} 
          & \xmark  &  \xmark    & 0.285 & 0.391 & 0.483 & 0.539 && 0.138 & 0.208 & 0.299 & 0.365 & 2.708 \\ 
        & & \cmark  &  \xmark    & \textbf{0.345} & \textbf{0.436} & \textbf{0.528} & \textbf{0.572} && \textbf{0.169} & \textbf{0.259} & \textbf{0.357} & \textbf{0.425} & \textbf{3.091} \\ [1pt]
        \cmidrule(r){5-14} 
        & & \xmark  &  \cmark    & 0.297 & 0.403 & 0.494 & 0.556 && 0.143 & 0.216 & 0.314 & 0.379 & 2.802 \\
        & & \cmark  &  \cmark    & 0.332 & 0.424 & 0.510 & 0.558 && 0.154 & 0.234 & 0.333 & 0.399 & 2.944 \\
        \cmidrule(r){2-14} 
        &  \multirow{2}{*}{ C3D}      
        &  - & \xmark       & 0.269 & 0.362 & 0.468 & 0.544    &&  0.130 & 0.198 & 0.288 & 0.352 & 2.611 \\
        & & -  &  \cmark    & 0.278 & 0.358 & 0.477 & 0.538    &&  0.133 & 0.200 & 0.289 & 0.357 & 2.630 \\
 \midrule
    \multirow{6}{*}{\textbf{Movies}} &     \multirow{4}{*}{Inception-v3} 
          & \xmark  & \xmark     & 0.187 & 0.254 & 0.341 & 0.407 && 0.118 & 0.173 & 0.248 & 0.302 & 2.030 \\ 
        & & \cmark  &  \xmark    & \textbf{0.217} & \textbf{0.289} & \textbf{0.375} & 0.428 && \textbf{0.136} & \textbf{0.199} & \textbf{0.277} & \textbf{0.333} & \textbf{2.254} \\ [1pt]
        \cmidrule(r){5-14} 
        & & \xmark  &  \cmark    & 0.216 & 0.263 & 0.354 & 0.413 && 0.123 & 0.180 & 0.254 & 0.306 & 2.109 \\
        & & \cmark  &  \cmark    & 0.214 & 0.279 & \textbf{0.375} & \textbf{0.437} && 0.134 & 0.191 & 0.265 & 0.320 & 2.215 \\
        \cmidrule(r){2-14} 
        &  \multirow{2}{*}{ C3D}      
        & -      & \xmark  & 0.180 & 0.251 & 0.332 & 0.392    &&  0.116 & 0.164 & 0.231 & 0.283 & 1.949 \\
        & & -    & \cmark  & 0.194 & 0.253 & 0.330 & 0.391    &&  0.118 & 0.167 & 0.234 & 0.287 & 1.974 \\

\bottomrule
\end{tabular}
 }
\end{table*}

%% file: table-video-da.tex
\begin{table} [tb!]
\renewcommand{\arraystretch}{1.2}
\caption{The effect of $\mu$ and $\sigma$ for video-level feature augmentation. Feature: Inception-v3.
}
\label{tab:perf-video-da}
\centering
 \scalebox{1.0}{
     \begin{tabular}{@{} lcc*{2}{c} @{}}
\toprule
\multirow{2}{*}{}    && \multicolumn{2}{c}{\textbf{Dataset}} \\
\cmidrule(r){2-4}
    && TV-shows   & Movies \\
\midrule
    $\mu=0, \sigma=0.1$                     && 2.705 &  2.021  \\
    $\mu=0, \sigma=1$                       && 2.667 &  2.070  \\
    $\mu=0, \sigma=10$                      && 2.654 &  1.982  \\
    estimated $\mu$ and $\sigma$            && \textbf{2.798} &  \textbf{2.109}  \\
\bottomrule
\end{tabular}
 }
\end{table}

%% file: sota.tex
\input{table-sota}

\subsection{Experiment 5: Strategy 1 vs. Strategy 2}

\begin{figure}[tb!]
\centering
 \subfigure [TV-shows] {
\noindent\includegraphics[width=0.47\columnwidth]{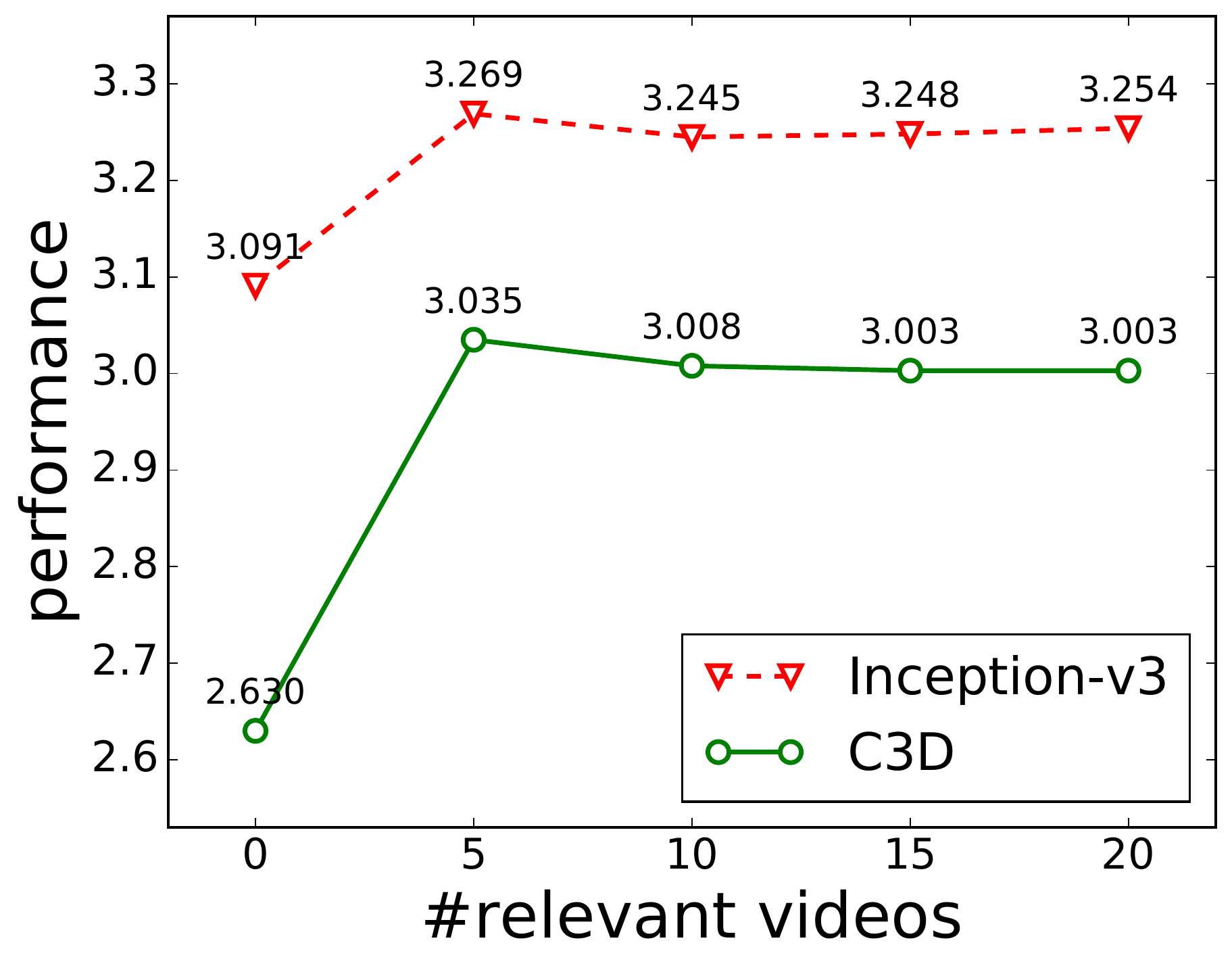}
\label{fig:relevant_video_k_shows}}
 \subfigure[Movies] {
\noindent\includegraphics[width=0.47\columnwidth]{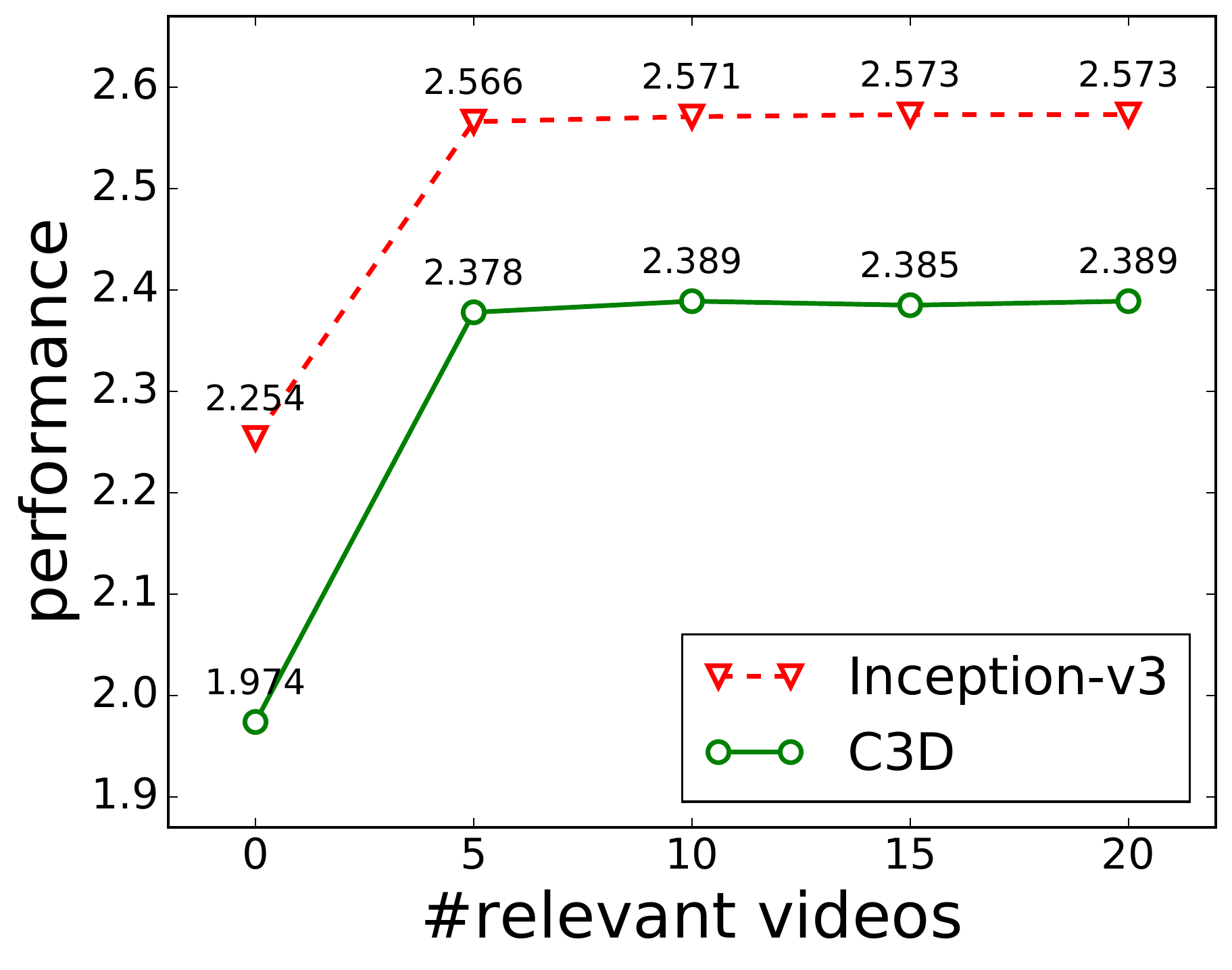}
\label{fig:relevant_video_k_movies}}
\caption{strategy 1 vs. strategy 2 for video relevance prediction on the (a) TV-shows and (b) Movies datasets.}
 \label{fig:relevant_video_k}
\end{figure}

In this experiment, we compare strategy 1 and strategy 2 for video relevance prediction in the re-learned video feature space.
The results on TV-shows and Movies are shown in Figure \ref{fig:relevant_video_k}.
It is worth noting that the starting point indicates strategy 1, while the others utilize strategy 2 with varying top $n$ relevant videos used.
It is clear that strategy 2 outperform strategy 1 with a large margin, which demonstrates that additionally using the relationship of a candidate video to another candidate video is helpful if available.
Furthermore, we observe that strategy 2 achieves comparable performance at the varied $n$ of 5, 10, 15 and 20. This observation basically verifies that strategy 2 has a good property of being not sensitive to the change of the number of the top relevant videos employed.

\subsection{Experiment 6: Comparison to the State-of-the-Art}\label{ssec:sota}

We compare our model with the state-of-the-art.
For better performance, our model used in this section is trained with frame-level feature augmentation strategy with multiple skip samplings of 8, 10 and 12 together.

\subsubsection{Comparison on the validation set}
For a fair comparison, we consider two scenarios.
For scenario 1, we assume that the relationship of a candidate video to another candidate video is unavailable, so we use strategy 1 to predict the video relevance. For scenario 2, we assume that the relationship is known, so strategy 2 is employed for our solution.
Table \ref{tab:perf-on-val-new} shows the performance of different models in different scenarios. 
In scenario 1, our proposed solution obtains the best overall performance. Note that \cite{chen2018content} reports only hit@30 and recall@100. This competitor has a higher recall@100, \ie 0.277 versus 0.262 on TV-shows and 0.202 versus 0.201 on Movies. Our model scores hit@30 with a larger margin, \ie 0.578 versus 0.491 on TV-shows and 0.439 versus 0.372 on Movies.
Other works such as \cite{hulu-baseline,fusedlstm} use TRL, the performance of which is inferior to ours. 
Compared to our conference results \cite{ourmm18}, the new solution obtains relative improvement of 4.8\% and 4.4\% in terms of the overall performance on TV-shows and Movies, respectively.

In scenario 2, our method again outperforms the state-of-the-art \cite{chen2018content}.
Notice that \cite{chen2018content} is a strong baseline as it employs model ensemble for better performance. Still, our single model is better.
The relative improvement of our model over this baseline is approximately 10\% and 15\% on TV-shows and Movies, respectively.
Note that all the works being compared are provided with the same features by the task organizers. So these results allow us to conclude the viability of the proposed feature re-learning method.

\subsubsection{Comparison on the test set}\label{ssec:leaderbord}

\begin{figure}[tb!]
\centering
 \subfigure [TV-shows] {
\noindent\includegraphics[width=0.9\columnwidth]{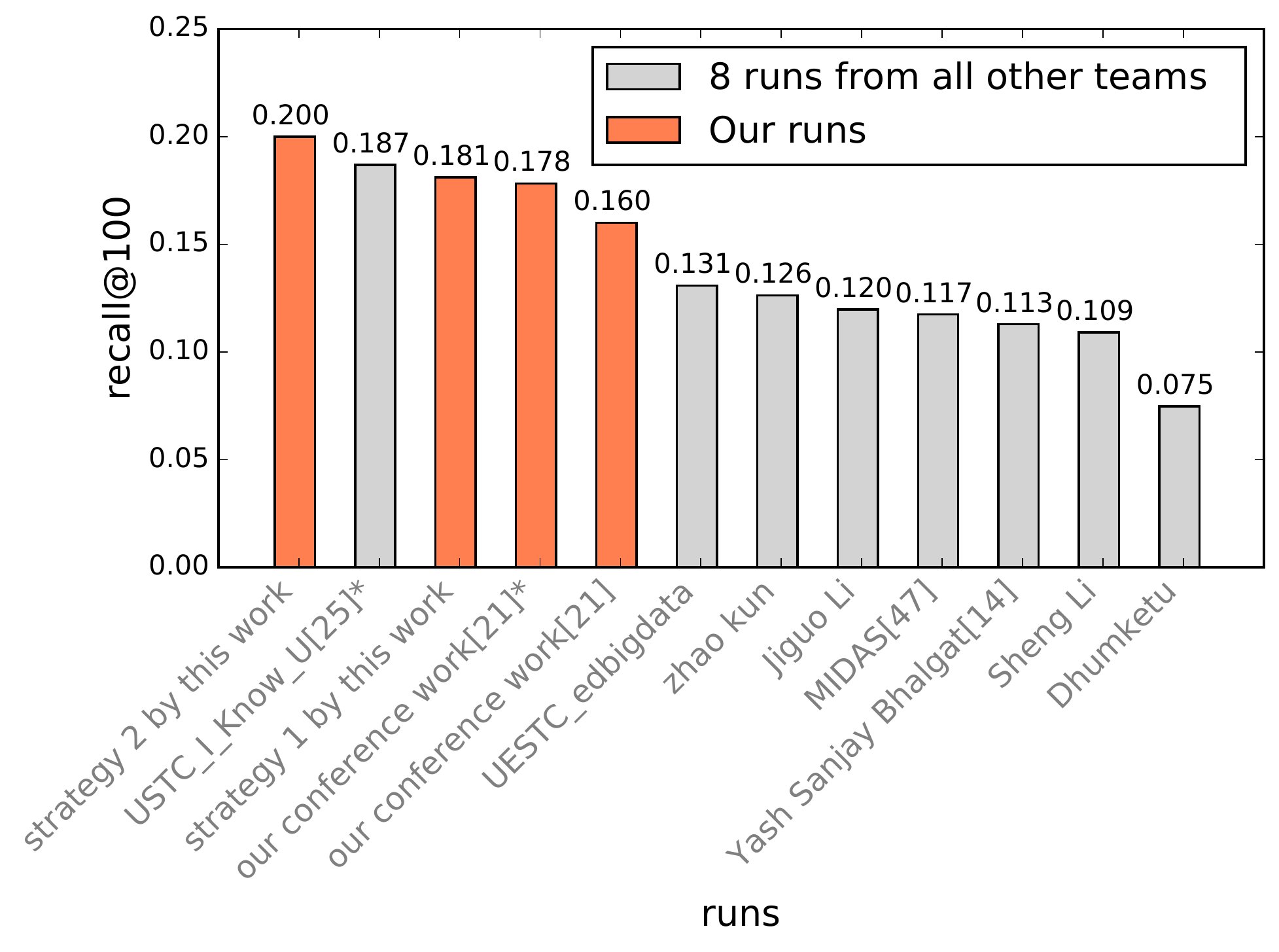}
\label{fig:robust_shows}}
 \subfigure[Movies] {
\noindent\includegraphics[width=0.9\columnwidth]{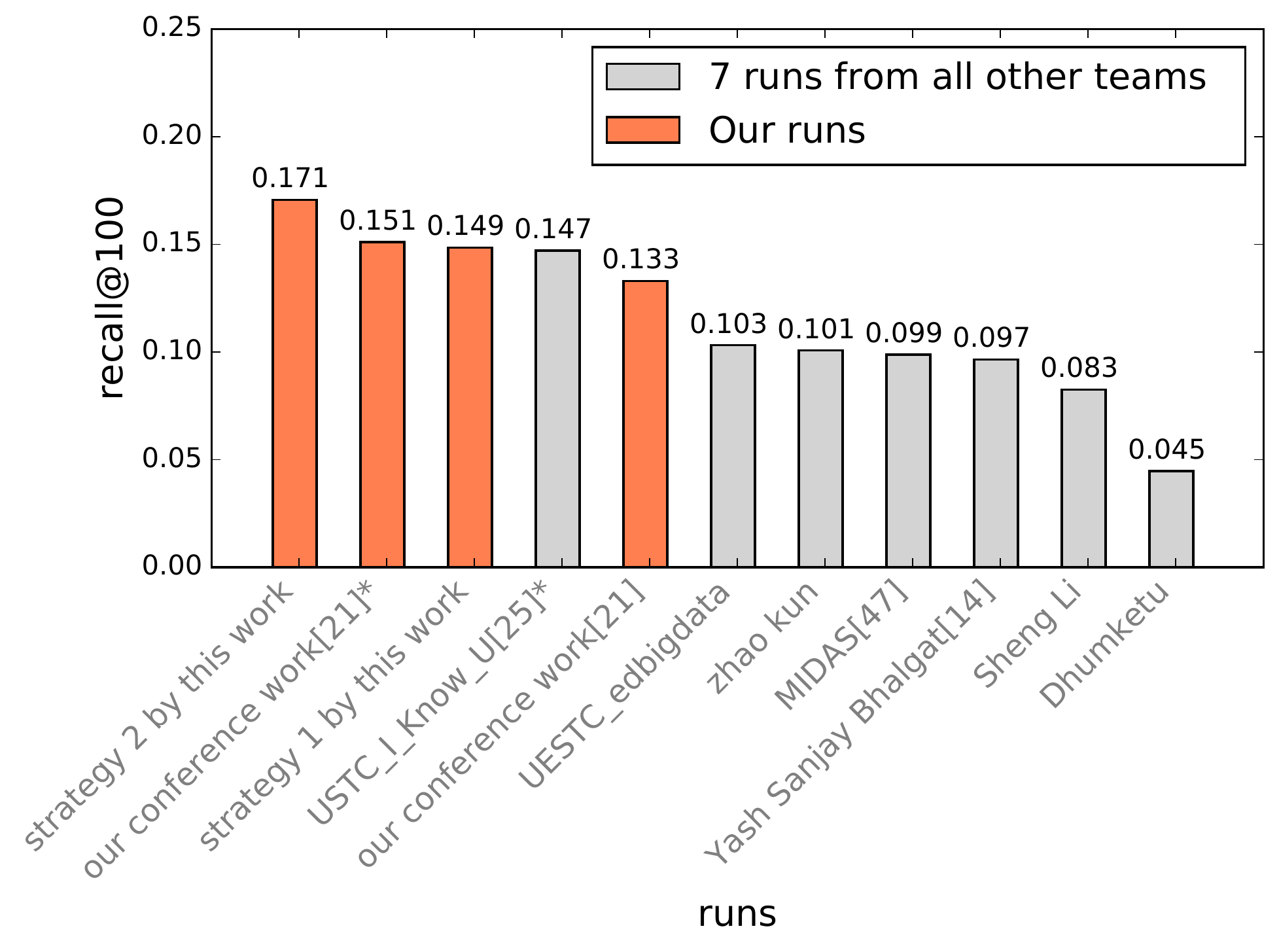}
\label{fig:robust_movies}}
\caption{State-of-the-art for video relevance prediction on the test of TV-shows and Movies respectively, showing the good performance of our solution compared to alternative approaches evaluated by the challenge organizers, where methods marked with * utilize the model ensemble strategy.
}
 \label{fig:leaderboard}
\end{figure}

The above evaluations are all conducted on the validation set, but the public availability of the ground-truth labels may unconsciously increase the chance of over-fitting. Hence, we further include in our evaluation on the test set, where the ground-truth is non-public and the performances are evaluated by the HULU challenge organizers.
As shown in Figure \ref{fig:leaderboard}, our runs lead the evaluation, which again verifies the effectiveness of our proposed method. 
Compared with our conference work, the models by this work give superiority performance, even outperform the conference run with the model ensemble. 
In the same setting without the model ensemble, our strategy 1 of video relevance prediction gives relative improvements of 13.1\% 12.0\% over the conference run on the TV-shows and Movies datasets respectively.
Moreover, the strategy 1 can be further improved by additionally using the relationship of a candidate video to another candidate video, that is strategy 2, lifting the performance from 0.181 to 0.200 on TV-shows and from 0.149 to 0.171 on Movies respectively. The results again show the benefit of using the relationship to predict video relevance.

\subsection{Efficiency Analysis}\label{ssec:efficiency}
Once our model is trained, given a pair of videos with extracted Inception-v3 features, it takes approximately 0.5 millisecond to predict video relevance. The speed is fast enough for video relevance based applications.
The performance is tested on a normal computer with 32G RAM and a GTX 1080TI GPU.

%% file: table-sota.tex
\begin{table*} [tb!]
\renewcommand{\arraystretch}{1.2}
\caption{Performance of the state-of-the-art on the validation set. Note that \cite{hulu-baseline,fusedlstm,kumar2018icebreaker,ourmm18} do not consider any relationship between candidate videos, so they are placed in Scenario 1.
}
\label{tab:perf-on-val-new}
\centering
 \scalebox{1.0}{
     \begin{tabular}{@{} lll*{10}{c} @{}}
\toprule
\multirow{2}{*}{\textbf{Dataset}} & \multirow{2}{*}{\textbf{Scenario}}  &  \multirow{2}{*}{\textbf{Method}} & \multicolumn{4}{c}{\textbf{hit@k}} &  & \multicolumn{4}{c}{\textbf{recall@k}} & \multirow{2}{*}{\textbf{Sum}} \\
         \cmidrule(r){4-7}  \cmidrule(r){9-12} 
    &  &  & k=5 & k=10 & k=20 & k=30 & &  k=50 & k=100 & k=200 & k=300 & \\
\midrule
    \multirow{7}{*}{\textbf{TV-shows}} 
        & \multirow{4}{*}{\textbf{\textit{Scenario 1:}}}
        & Liu \etal \cite{hulu-baseline}         & 0.247 & 0.338 & 0.450 & 0.530 && 0.111 & 0.172 & 0.262 & 0.331 & 2.441 \\
        && Bhalgat \etal \cite{fusedlstm}         & 0.265 & 0.343 & 0.435 & 0.483 && 0.139 & 0.205 & 0.277 & 0.327 & 2.474 \\
        && Kumar \etal \cite{kumar2018icebreaker}       & 0.319 & 0.388 & 0.485 & 0.541 && 0.171 & 0.229 & 0.289 & 0.323 & 2.745 \\
        && Dong \etal \cite{ourmm18}       & 0.315 & 0.409 & 0.510 & 0.558 && 0.162 & 0.244 & 0.344 & 0.408 & 2.950 \\
        && Chen \etal \cite{chen2018content}      & - & - & - & 0.491 &&  - & \textbf{0.277} & - & - & - \\
        && this work         & \textbf{0.334} & \textbf{0.433} & \textbf{0.527} & \textbf{0.578} && \textbf{0.173} & 0.262 & \textbf{0.359} & \textbf{0.426} & \textbf{3.092} \\
        \cmidrule(r){2-13}
        & \multirow{2}{*}{\textbf{\textit{Scenario 2:}}}
        & Chen \etal \cite{chen2018content}      & - & - & - & 0.528 &&  - & 0.295 & - & - & - \\
        && this work        & \textbf{0.350} & \textbf{0.436} & \textbf{0.538} & \textbf{0.581} && \textbf{0.201} & \textbf{0.296} & \textbf{0.399} & \textbf{0.468} & \textbf{3.269} \\
 \midrule
    \multirow{7}{*}{\textbf{Movies}} 
        &\multirow{4}{*}{\textbf{\textit{Scenario 1:}}}
        & Liu \etal \cite{hulu-baseline}         & 0.133 & 0.187 & 0.269 & 0.312 && 0.086 & 0.125 & 0.185 & 0.229 & 1.526 \\
        && Bhalgat \etal \cite{fusedlstm}         & 0.165 & 0.193 & 0.315 & 0.383 && 0.112 & 0.173 & 0.207 & 0.281 & 1.829 \\
        && Kumar \etal \cite{kumar2018icebreaker}       & 0.180 & 0.231 & 0.313 & 0.362 && 0.099 & 0.139 & 0.192 & 0.226 & 1.742 \\
        && Dong \etal \cite{ourmm18}       & 0.196 & 0.274 & 0.363 & 0.430 && 0.132 & 0.191 & 0.274 & 0.333 & 2.193 \\
        && Chen \etal \cite{chen2018content}      & - & - & - & 0.372 &&  - & \textbf{0.202} & - & - & - \\
        && this work         & \textbf{0.227} & \textbf{0.293} & \textbf{0.377} & \textbf{0.439} && \textbf{0.142} & 0.201 & \textbf{0.276} & \textbf{0.334} & \textbf{2.289} \\
        \cmidrule(r){2-13}
        &\multirow{2}{*}{\textbf{\textit{Scenario 2:}}} 
        & Chen \etal \cite{chen2018content}      & - & - & - & 0.402 &&  - & 0.232 & - & - & - \\
        && this work        & \textbf{0.234} & \textbf{0.307} & \textbf{0.410} & \textbf{0.476} && \textbf{0.167} & \textbf{0.245} & \textbf{0.333} & \textbf{0.394} & \textbf{2.566} \\

\bottomrule
\end{tabular}
 }
\end{table*}